\documentclass[10pt]{article} 
\usepackage[preprint]{tmlr}

\usepackage{amsmath,amsfonts,bm}

\def\eqref#1{equation~\ref{#1}}

\def\1{\bm{1}}

\DeclareMathAlphabet{\mathsfit}{\encodingdefault}{\sfdefault}{m}{sl}
\SetMathAlphabet{\mathsfit}{bold}{\encodingdefault}{\sfdefault}{bx}{n}

\usepackage{hyperref}
\usepackage{url}
\usepackage{booktabs}
\usepackage{graphicx}
\usepackage{xcolor}
\usepackage{subcaption}
\usepackage{relsize}
\title{\textsc{AEyeDE}: An Attention-Based Attribution Framework for AI-Generated Text Detection}

\author{\name Aria Nourbakhsh\thanks{Both authors contributed equally.} \email aria.nourbakhsh@uni.lu \\
      \addr Department of Computer Science\\
      University of Luxembourg
      \AND
      \name Adelaide Danilov\footnotemark[1] \email adelaide.danilov.002@student.uni.lu \\
      \addr Department of Computer Science\\
      University of Luxembourg
      \AND
      \name Christoph Schommer \email christoph.schommer@uni.lu\\
      \addr Department of Computer Science\\
      University of Luxembourg
      \AND
      \name Salima Lamsiyah \email salima.lamsiyah@uni.lu\\
      \addr Department of Computer Science\\
      University of Luxembourg}

\def\month{MM}  
\def\year{YYYY} 
\def\openreview{\url{https://openreview.net/forum?id=XXXX}} 

\begin{document}

\maketitle

\begin{abstract}
Detecting AI-generated text is becoming increasingly challenging as modern language models approach human-level fluency and can evade detectors that rely on surface statistics or likelihood-based signals. We propose \textsc{AEyeDE}, an attribution-driven approach to human-AI authorship detection that leverages model attention as a discriminative signal. Specifically, we extract attention-based attribution matrices for both human- and AI-generated text using a \emph{proxy} Transformer model with white-box access and train a lightweight Convolutional Neural Network to learn representations from these attribution maps. Across encoder-decoder translation settings, our method consistently outperforms a text-only baseline. In decoder-only settings, it performs strongly in generator-specific detection, remains competitive on standard benchmarks, and shows robustness under cross-dataset transfer and alternative-spelling perturbations. We further show that attention maps exhibit recurring local structures whose relative frequencies differ consistently between human- and AI-generated text across datasets and proxy models. These findings suggest that attention-based attribution maps provide a complementary and interpretable signal for AI-generated text detection. We will make the code publicly available to support future research.

\end{abstract}

\section{Introduction}
The advent of large language models (LLMs) has enabled the generation of coherent, context-aware, and human-like text across a wide range of domains and languages \citep{naveed2023comprehensive,chang2024survey}. While these advances unlock substantial benefits, they also raise critical challenges related to information integrity, authorship, and misuse, including large-scale misinformation in journalism and academic dishonesty in educational settings, where automated content generation threatens societal trust, originality, and assessment validity \citep{dugan2023real,wu2025survey,liu2024detectability,ali-etal-2025-detection,HUANG2025107877,bittle2025generative}.

In response, several AI-generated text detection methods have been proposed, including surface-statistical approaches exploiting cues such as perplexity, burstiness, and n-gram repetition \citep{gehrmann-etal-2019-gltr,ippolito-etal-2020-automatic}; likelihood-based methods that probe variation in a model’s probability landscape via perturbation or re-sampling \citep{mitchell2023detectgpt,bao2023fast}; supervised classifiers that fine-tune Transformer encoders on labeled data \citep{li-etal-2025-iron,zhu-etal-2025-reliably}; and watermarking techniques that embed detectable signals for source attribution \citep{kirchenbauer2023watermark,liu2024survey}. However, each paradigm has intrinsic limitations: statistical and likelihood-based detectors degrade as LLMs are optimized to mimic human distributions through techniques such as RLHF \citep{christiano2017deep}; supervised classifiers suffer under domain shift and unseen generators \citep{uchendu2021turingbench}; and watermarking requires model-side cooperation and is fragile to paraphrasing, post-editing, or partial reuse \citep{liu2024survey,wang-etal-2025-morphmark,niess2025ensemble,ahn2025ditto}. As a result, robust detection remains an open challenge, continually undermined by advances in generation quality and adversarial evasion strategies \citep{wu2025survey,wu2024detectrl}.

These challenges motivate a shift away from detecting \emph{what} is written toward analyzing \emph{how} text is produced. 
In particular, we hypothesize that token-token interaction structure provides a richer detection signal than one-dimensional token-level statistics such as rank, likelihood, or perplexity. Text-only detectors, whether statistical or neural, operate on the generated sequence itself. Attention-based attribution maps, by contrast, capture how a proxy Transformer internally processes that sequence, revealing a two-dimensional relational structure over token interactions. This representation can preserve higher-order local regularities that are not directly observable from surface-form statistics alone. We therefore investigate whether these attribution patterns provide a more robust and transferable signal for distinguishing human- and AI-generated text. To test this hypothesis, we introduce \textsc{AEyeDE}, an attribution-based detection framework that operates directly on attention attribution maps extracted from Transformer models \citep{10.5555/3295222.3295349}.  Given an observed text $x$ (human or AI-generated), \textsc{AEyeDE} passes $x$ through a fixed \emph{proxy} model $G_\theta$ with white-box access and derives an attention attribution matrix (Sec.~\ref{sec:framework}). We process attribution maps using a multi-scale convolutional encoder with attention pooling to obtain compact embeddings for authorship classification (Figure~\ref{fig:architecture}), making the detector less sensitive to purely lexical or stylistic variation.

Beyond predictive performance, this approach also enables an interpretable analysis of the learned attribution structure. For this purpose, we analyze what the CNN attribution encoder captures in attention maps. 
Clustering $8\times8$ patches in its last convolutional stage feature space reveals recurring local patterns (\emph{motifs}) whose prevalence differs between human and AI-generated text across datasets and proxy models. 
This indicates that authorship leaves a localized, repeatable signature in proxy-model attention maps that our detector can exploit.

We evaluate \textsc{AEyeDE} across both encoder-decoder and decoder-only settings, using machine translation benchmarks (WMT14 and the UN Parallel Corpus) and open-ended generation datasets (\textsc{HC3}, \textsc{RAID}, and Beemo). Together, these experiments cover multiple languages, domains, and model families.

Our main contributions are summarized as follows:
\begin{itemize}
\item We introduce \textsc{AEyeDE}, an attribution-based framework for AI-generated text detection that uses attention attribution maps from a proxy Transformer as structured input to a lightweight CNN classifier.

\item We provide a broad empirical evaluation across encoder-decoder and decoder-only settings, including generator-specific detection, mixed-generator generalization, adversarial perturbations, and cross-dataset transfer. The results show that attention-based attribution maps provide a competitive and complementary detection signal, with particularly strong performance in generator-specific settings and robustness to alternative-spelling attacks.

\item We analyze the learned attribution representations and identify recurring local attention patterns, or ``motifs,'' whose relative frequencies differ systematically between human- and AI-generated text. These motifs provide an interpretable and localized signature of authorship.

\end{itemize}

\section{Related Work}

\paragraph{AI-Generated Text Detection.}
Research on detecting machine-generated text has accelerated alongside the rapid progress and deployment of LLMs \citep{wu2025survey}. 
Existing approaches can be categorized into (i) \emph{surface-statistical} and (ii) \emph{likelihood-based} detectors, (iii) \emph{supervised neural classifiers}, (iv) \emph{watermarking and source attribution}, and (v) \emph{LLM-based} meta-detectors. 
Surface-statistical methods exploit distributional artifacts such as perplexity, burstiness, or n-gram irregularities, often providing lightweight but increasingly fragile signals as generators improve \citep{gehrmann-etal-2019-gltr,ippolito-etal-2020-automatic,shen2023textdefense,tassopoulou2021enhancing,krishna-etal-2022-rankgen}. 
Complementarily, likelihood-based methods probe the generator’s probability landscape: DetectGPT identifies machine-generated text by measuring curvature via perturbations \citep{mitchell2023detectgpt}, and related work improves efficiency and robustness through faster perturbation schemes \citep{bao2023fast}.  These lines of work capture model-specific statistical footprints, but can degrade as LLMs are optimized to match human-like distributions.

\paragraph{Neural Detectors, Robustness, and Generalization.}
Supervised detectors typically fine-tune Transformer encoders (e.g., BERT \citep{devlin-etal-2019-bert} and RoBERTa \citep{liu2019roberta}) on labeled human vs.\ machine text, achieving strong in-domain performance but often suffering under domain shift and unseen generators \citep{uchendu2021turingbench,wang-etal-2024-m4}. 
Robustness has become a central focus: training on diverse decoding strategies improves resilience \citep{ippolito-etal-2020-automatic}, adversarial training frameworks such as IRON harden detectors against evasion \citep{li-etal-2025-iron}, and Radar explicitly targets robustness via adversarial learning \citep{hu2023radar}. 
Recent methods further aim to improve out-of-distribution behavior and reliability guarantees, e.g., by shaping attention over multiple receptive ranges \citep{jiao-etal-2025-rangedetector} or bounding false positives with conformal prediction in zero-shot settings \citep{zhu-etal-2025-reliably}. 
Interpretability for detectors is also receiving attention: feature-level analyses using sparse autoencoders help reveal which latent patterns separate machine and human text \citep{kuznetsov-etal-2025-feature}, while downstream applications increasingly require multilingual and domain-specific robustness \citep{ali-etal-2025-detection} and fine-grained settings such as human-AI co-authorship \citep{su-etal-2025-haco}.

\paragraph{Watermarking and Source Attribution.}
Watermarking aims to embed detectable signals into generated text, enabling attribution when generation-side cooperation is available \citep{liu2024survey}. 
Early and widely adopted schemes include token-list or ``soft'' watermarks that bias sampling \citep{kirchenbauer2023watermark}, while subsequent work explored alternative embedding mechanisms and detection rules, including entropy- or Bayesian-inspired detectors \citep{lu-etal-2024-entropy,huang2025low} and more adaptive watermark designs such as MorphMark \citep{wang-etal-2025-morphmark}. 
Recent studies further examine watermark ensembles \citep{niess2025ensemble}, watermark-based source attribution (e.g., WASA) \citep{lu2025wasa}, and approaches that reduce bias and risk \citep{mao2025watermarking}. 
However, watermarking remains challenged by post-editing and paraphrasing \citep{liu2024survey}, motivating defenses such as paraphrase inversion \citep{rivera-soto-etal-2025-mitigating} and robustness through injected ``fictitious knowledge'' signals \citep{cui2025robust}. 
Adversarial settings also reveal vulnerabilities: DITTO formalizes spoofing attacks against watermarked LLMs via knowledge distillation \citep{ahn2025ditto}, underscoring the need for evaluation under realistic transformation and attack pipelines.

\paragraph{LLMs as Detectors and Explainable Attribution.}
Beyond classical detectors, LLMs are increasingly used as meta-detectors and critics of generated content, reflecting a trend toward black-box and instruction-following detection pipelines \citep{wang-etal-2024-m4}. 
Recent work expands from binary detection to attribution and explanation, e.g., XDAC provides XAI-driven detection and attribution for Korean news comments \citep{go-etal-2025-xdac,go2025xdac}, and studies of detectability highlight how author intent and role can affect detection outcomes \citep{li-wan-2025-writes}. 
Together, these directions emphasize that practical detection increasingly requires robustness, reliability, and interpretable evidence—not only raw accuracy.

\paragraph{Benchmarks and Shared Tasks.}
Progress in AI-text detection is tightly coupled with benchmarks that stress generalization across domains, languages, and attack conditions. 
Widely used datasets include HC3 \citep{guo2023close}, MGTBench \citep{he2024mgtbench}, WritingPrompts \citep{bao2023fast}, RAID \citep{dugan-etal-2024-raid}, and adversarial extensions such as Adv-HC3 \citep{peng2023hidding}; additional resources target broader settings such as BUST \citep{cornelius2024bust} and LLMTRACE \citep{tolstykh2025llmtrace}. 
Beyond text-only benchmarks, MultiSocial supports multilingual social-media detection \citep{macko-etal-2025-multisocial}, Double Entendre introduces a multimodal audio-lyrics setting \citep{frohmann-etal-2025-double}, and stress-test benchmarks systematically perturb style to probe brittleness \citep{pedrotti-etal-2025-stress}. 
Shared tasks further standardize evaluation and accelerate methodology: SemEval-2024 Task 8 targets black-box, multilingual, and multidomain detection \citep{wang-etal-2024-semeval-2024}, with system analyses such as TrustAI highlighting practical modeling choices \citep{urlana-etal-2024-trustai}. 
Community efforts such as the GenAIDetect workshop at COLING 2025 \citep{genaidetect-ws-2025-1} and domain-focused shared tasks and datasets (e.g., M-DAIGT for news and academic writing \citep{lamsiyah2025m}, and AraGenEval for Arabic settings \citep{abudalfa2025arageneval}) reflect increasing emphasis on robustness, multilinguality, and real-world constraints. 
These benchmarks and tasks collectively motivate detectors that generalize across generator families while offering transparent, verifiable evidence for their decisions.

\paragraph{Positioning of Our Work.} Our method differs from prior AI-text detectors by using attention-based attribution maps from a proxy Transformer as the main detection signal, rather than relying only on surface statistics, likelihood perturbations, or end-to-end text representations. It does not require watermarking or access to the true generator, and it is less dependent on lexical or stylistic cues alone. More specifically, our framework uses attribution maps as structured inputs to a dedicated detection model. The experiments suggest that this signal is useful for this task: it performs especially well in generator-specific settings, remains competitive on standard benchmarks, and shows robustness under alternative-spelling attacks and cross-dataset transfer.

\section{\textsc{AEyeDE} Framework}\label{sec:framework}

\begin{figure*}[t]
    \centering
    \includegraphics[scale=0.7]{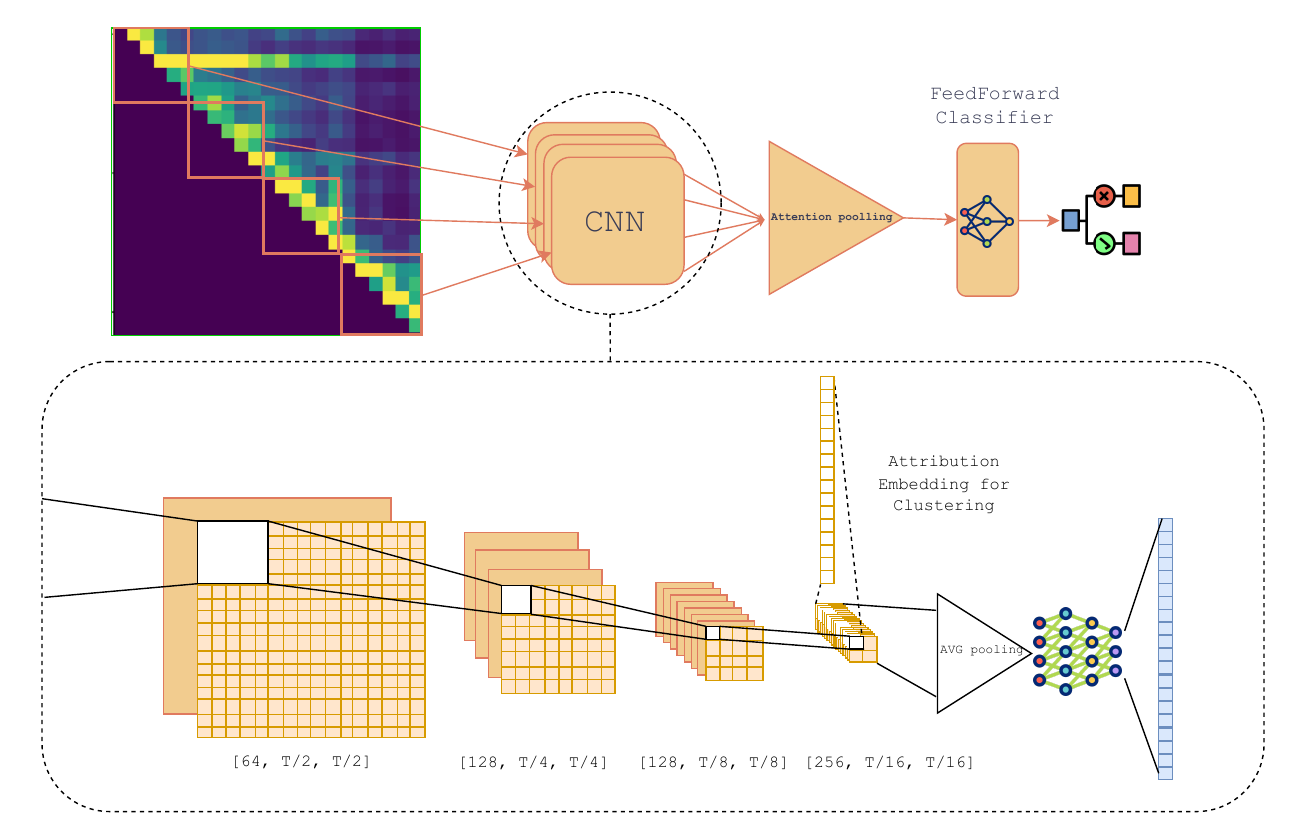}
    \caption{Overview of \textsc{AEyeDE}. Given a text sample and white-box access to a proxy generator model \(G_\theta\), we extract an attention-derived attribution matrix \(A\) (top left). For decoder-only models, we summarize \(A\) by sampling fixed-size square blocks (e.g., \(128\times128\)) along the main diagonal, where the strongest local token-token interactions are concentrated (orange boxes) \citep{xiao2023efficient,ivanitskiy2025motifs,qi2025deltallm}; for encoder-decoder models, we use the full cross-attention map. Each block is encoded by a CNN (bottom), producing one embedding per block. These embeddings are aggregated with learnable attention pooling to form a global attribution representation, which is passed to a lightweight MLP classifier to predict human versus LLM authorship (top right). In our implementation, the final low-resolution CNN feature map is \(16\times16\); its spatial cells correspond to regions of the original attribution block and are later used for motif clustering.}
    \label{fig:architecture}
\end{figure*}

We cast AI-text detection as a binary classification problem. The detector has white-box access to a proxy language model \(G_\theta\), which may be either the suspected generator itself or a surrogate model. Given an observed text sample \(x\), we pass \(x\) through \(G_\theta\) and extract an attention-derived attribution map. Within each experiment, the same proxy model is used to compute attributions for all inputs, regardless of whether \(x\) is human-written or AI-generated. Our hypothesis is that the internal processing dynamics of \(G_\theta\) induce systematic differences in these attribution maps for human versus machine-generated text.

For decoder-only models, the attribution map reflects the influence of previous tokens on each current token under causal self-attention. For encoder-decoder models, it reflects the influence of source tokens on each target token under cross-attention. In both cases, the attribution map serves as the primary input to the downstream detector.

Let \(\mathcal{V}\) denote the vocabulary, and let \(x=(x_1,\ldots,x_T)\in\mathcal{V}^T\) denote the text whose authorship is to be predicted. Let \(y\in\{0,1\}\) be the corresponding label, where \(y=1\) indicates AI-generated text and \(y=0\) indicates human-written text. When the proxy model is encoder-decoder, the target text \(x\) is paired with its associated source sequence to compute cross-attention. The detector is a conditional classifier
\[
D_\phi(x;\theta)=f_\phi\!\big(x, A(x;\theta)\big)\in[0,1],
\]
where \(A(x;\theta)\) is the attention-derived attribution map extracted from \(G_\theta\), and \(f_\phi\) is a learnable decision function with parameters \(\phi\).

Assume first that \(G_\theta\) is an \(L\)-layer, \(H\)-head Transformer. At layer \(\ell\in\{1,\ldots,L\}\) and head \(h\in\{1,\ldots,H\}\), let
\(Q^{(\ell,h)}\in\mathbb R^{T_t\times d_k}\) and
\(K^{(\ell,h)}\in\mathbb R^{T_s\times d_k}\)
denote the query and key matrices. The attention matrix for that head is
\[
\widetilde{A}^{(\ell,h)}(x;\theta)
=
\operatorname{softmax}\!\left(
\frac{Q^{(\ell,h)}(K^{(\ell,h)})^\top}{\sqrt{d_k}} + C
\right)
\in\mathbb R^{T_t\times T_s},
\]
where \(C\in\mathbb R^{T_t\times T_s}\) is the causal mask in the decoder-only setting. In the encoder-decoder setting, \(\widetilde{A}^{(\ell,h)}\) denotes cross-attention with the same target-by-source orientation and no causal mask. We then average across heads and layers to obtain a single attribution map:
\[
A(x;\theta)
=
\frac{1}{LH}
\sum_{\ell=1}^{L}\sum_{h=1}^{H}
\widetilde{A}^{(\ell,h)}(x;\theta)
\in\mathbb R^{T_t\times T_s}.
\]
Each entry
\[
a_{t,s}=\big[A(x;\theta)\big]_{t,s}
\]

gives the average attention mass assigned from the target position \(t\) to source/previous position \(s\) under the proxy model. In the decoder-only setting, \(T_s=T_t=T\); in the encoder-decoder setting, \(T_s\) and \(T_t\) may differ. We use the \texttt{Inseq} library to extract these attention-derived attributions~\citep{sarti-etal-2023-inseq}.

The detector architecture is shown in Fig.~\ref{fig:architecture}. For decoder-only models, we summarize \(A\) by extracting square blocks of size \(w_a\) along a diagonal traversal; in all experiments, we set \(w_a=128\). In encoder-decoder experiments, the inputs are limited to at most 128 tokens, so we use a single block that covers the full cross-attention map.\footnote{For example, on HC3 (see Section~\ref{subsec-data}), the \(128\times128\) diagonal band exhibits higher entropy than the off-diagonal region (8.89 vs.\ 7.78), together with a higher mean attribution value (0.007 vs.\ 0.001) and a larger standard deviation (0.01 vs.\ 0.002). Processing four \(128\times128\) patches is also considerably more efficient than processing a full \(1024\times1024\) matrix.} This design is motivated by the empirical observation that, in decoder-only models, the most informative attribution mass is concentrated near the main diagonal \citep{xiao2023efficient,ivanitskiy2025motifs,qi2025deltallm}.

Let \((p_k,q_k)\) denote the top-left corner of the \(k\)-th block. We define
\begin{equation}
A_k = A[p_k:p_k+w_a,\; q_k:q_k+w_a] \in \mathbb R^{w_a\times w_a}.
\end{equation}
Let \(m_t\in\{0,1\}^{T_t}\) and \(m_s\in\{0,1\}^{T_s}\) denote the target- and source-side padding masks, respectively, and define the validity mask
\begin{equation}
P = m_t m_s^\top \in \{0,1\}^{T_t\times T_s},
\end{equation}
which marks non-padding attribution entries. A block is considered valid if it contains at least one non-padding entry:
\begin{equation}
u_k
=
\mathbb I\!\left(
\sum_{i,j} P[p_k+i,\; q_k+j] > 0
\right)\in\{0,1\}.
\end{equation}
Each valid block is encoded by a CNN-based attribution encoder
\begin{equation}
b_k = E_{\text{attr}}(A_k)\in\mathbb R^{d_{\text{attr}}}.
\end{equation}

The attribution encoder \(E_{\text{attr}}\) treats \(A_k\) as a single-channel image. It applies a sequence of convolutional blocks with channel progression
\[
1\rightarrow 32\rightarrow 64\rightarrow 128\rightarrow 128\rightarrow 256,
\]
interleaved with \(2\times2\) max-pooling, followed by global average pooling and a two-layer MLP. Writing \(g(\cdot)\) for the convolutional pipeline plus global average pooling, we obtain
\begin{equation}
r_k = g(A_k)\in\mathbb R^{256},
\end{equation}
and then
\begin{equation}
b_k
=
W_2\,\rho\!\big(\mathrm{BN}(W_1 r_k + b_1)\big)+b_2,
\end{equation}
where \(\rho(\cdot)\) denotes ReLU. In our implementation, the final convolutional feature map is \(16\times16\); its spatial cells can be mapped back to local regions of the original attribution block for motif clustering (see Sec.~\ref{sec-disccusion}).

Optionally, we augment the attribution representation with a text representation.\footnote{We evaluate this text-augmented variant for encoder-decoder models. In practice, attention-only representations are consistently competitive with the text-augmented variant, so we emphasize the attention-based results in the main paper.} To handle long sequences, we pad the input to a multiple of a fixed window size \(w_t\) and partition it into
\[
N=\left\lceil \frac{T}{w_t}\right\rceil
\]
contiguous chunks:
\begin{equation}
x^{(i)} = x_{(i-1)w_t+1:iw_t},
\qquad i=1,\ldots,N.
\end{equation}
A text encoder \(E_{\text{text}}\) maps each chunk to a vector
\begin{equation}
c_i = E_{\text{text}}\!\big(x^{(i)}\big)\in\mathbb R^{d_{\text{text}}}.
\end{equation}
We ignore fully padded chunks via a validity indicator \(v_i\in\{0,1\}\).

We aggregate sets of per-block vectors using learnable attention pooling. Given vectors \(\{z_i\}_{i=1}^n\) and a corresponding validity mask \(\{m_i\}_{i=1}^n\), we define
\begin{align}
s_i &= \omega^\top \tanh(W_p z_i), \\
\alpha_i &= \frac{m_i\exp(s_i)}{\sum_j m_j\exp(s_j)}, \\
\mathrm{Pool}(\{z_i\},\{m_i\}) &= \sum_i \alpha_i z_i,
\end{align}
where \(\omega\) and \(W_p\) are learnable parameters. Applying this operator to the text and attribution branches gives
\begin{equation}
h_{\text{text}} = \mathrm{Pool}(\{c_i\},\{v_i\}),
\qquad
h_{\text{attr}} = \mathrm{Pool}(\{b_k\},\{u_k\}).
\end{equation}

The final representation is
\begin{equation}
h=
\begin{cases}
[\,h_{\text{text}};\,h_{\text{attr}}\,], & \text{if the text branch is used},\\[2mm]
h_{\text{attr}}, & \text{otherwise}.
\end{cases}
\end{equation}
A two-layer MLP then produces a scalar logit \(\ell\) and the corresponding probability of AI authorship:
\begin{equation}
\ell = w^\top \rho(W_h h + b_h) + b,
\end{equation}
\begin{equation}
p(y=1\mid x,A) = \sigma(\ell).
\end{equation}
We train the detector using binary cross-entropy on labeled examples.

In this work, prompts are discarded in the decoder-only setting. Thus, the detector conditions only on the observed continuation \(x_{1:T}\) and the attribution map derived from it, rather than on the full prompt-response pair.

\section{Experimental Results}

\subsection{Datasets}
\label{subsec-data}
In this study, we evaluate both encoder–decoder and decoder-only language models. For the encoder–decoder setting, we conduct experiments on three translation language pairs: French–English (fr-en) and German–English (de-en) using WMT14 \citep{bojar2014findings}, and Arabic–English (ar-en) using the UN Parallel Corpus \citep{ziemski-etal-2016-united}. For each language pair, we construct a dataset of 200k source–target examples consisting of gold (human) reference translations and corresponding model-generated translations produced by the Marian-MT model~\citep{tiedemann-thottingal-2020-opus}. We selected samples with source and target pairs of at most 128 tokens. 

For the decoder-only setting, we use the \textsc{HC3} dataset \citep{guo2023close}, RAID \citep{dugan-etal-2024-raid}, and Beemo \citep{artemova-etal-2025-beemo}. \textsc{HC3} provides paired human and ChatGPT~\citep{openai2023chatgpt} responses. We remove examples exceeding 1,024 tokens and retain approximately 24k samples per class. RAID contains human-written text and model-generated text spanning multiple domains. From RAID, we use outputs from Cohere, Llama 2 70B (chat), GPT-2 XL, and Mistral-7B, together with the corresponding human responses to the same prompts. Because RAID is imbalanced (with fewer human than model-generated examples), we downsample each model’s generated subset to 26,700 examples (the minimum across selected models) and use all available human examples (12,900). We use Beemo as an external test dataset for cross-dataset out-of-distribution experiments. With our preprocessing on Beemo, we obtain 2,178 human-written samples and an equal number of machine-generated samples, which are sourced from multiple LLMs. In all experiments, we use a class-balanced entropy loss to mitigate the imbalance between human- and AI-generated samples.

We split each dataset into 90\% training, 5\% validation, and 5\% test. Throughout our experiments with decoder-only models, we discard prompts, as in practice, it is more likely that a text sample is observed without access to the prompt that generated it. We use \textsc{Llama} 3.1 8B \citep{grattafiori2024llama3herdmodels}, \textsc{Cohere} (c4ai-command-r7b-12-2024)\citep{cohere2025commandaenterprisereadylarge}, \textsc{GPT-neo}~\citep{GPT-neo}, and \textsc{Mistral} (Ministral-3-8B-Instruct-2512)~\citep{jiang2023mistral7b} to obtain the attributions from approximate models of the same family as those that generated the \textsc{RAID} and \textsc{HC3} datasets.

\subsection{Evaluation Metrics}

For evaluation, we report Accuracy, Precision, Recall, F1, Area Under Curve, and True/False-positive Rate at a fixed low false-positive operating point, namely
$\mathrm{TPR}@\mathrm{FPR}=0.01$.
Here $\mathrm{TPR}=\frac{\mathrm{TP}}{\mathrm{TP}+\mathrm{FN}}$ and
$\mathrm{FPR}=\frac{\mathrm{FP}}{\mathrm{FP}+\mathrm{TN}}$; thus
$\mathrm{FPR}=0.01$ corresponds to falsely labeling $1\%$ of human-written samples as AI-generated. This is the critical ``high-precision'' regime. In academic or forensic settings, false positives (accusing a human of using AI) are unacceptable. A detector must have a high TPR at a very low FPR to be deployable \citep{ayoobi2025beyond}. For all threshold-dependent metrics, thresholds were selected on the validation split and then fixed for test evaluation.

\subsection{Baseline Models}
For the encoder-decoder setting, our baseline is a custom three-layer Transformer-based classifier trained on paired source-target text. For the decoder-only setting, we report results from Fast-DetectGPT (Curvature)~\citep{bao2023fast}, Binoculars~\citep{hans2024spottingllmsbinocularszeroshot}, Rank and LogRank~\citep{su-etal-2023-detectllm}, GLTR~\citep{gehrmann-etal-2019-gltr}, and a RoBERTa-based detector released by SuperAnnotate.\footnote{\url{https://huggingface.co/SuperAnnotate/ai-detector}}

Curvature compares the conditional log-probability of each observed token with that of plausible alternative tokens in the same context, and aggregates these comparisons into a sample-level statistic called conditional probability curvature. Machine-generated text tends to show a higher curvature value because its token choices are more often locally optimal under the model. Binoculars computes the ratio between a model's own perplexity and its cross-perplexity with another model. The intuition is that machine-generated text is not only low-perplexity for an LLM, but also shows unusually high agreement across related models\footnote{We use LLama-3.1-8B and Llama-3.1-8 B-Instruct the pair models.}. GLTR detects machine-generated text by scoring each token under a reference language model using its probability, rank, and predictive entropy. Rank is a token-level measure that extracts the rank of each token under a language model and averages it across the sequence: lower Rank values indicate that the sample is more likely to be machine-generated. LogRank computes the logarithm of this rank-based score.

\subsection{Results}
We evaluate the proposed framework across five experimental settings, followed by an analysis of attribution-region motifs, each of which tries to answer a research question of the use of attention attribution for this task. First, we apply \textsc{AEyeDE} to encoder-decoder attention maps and examine the effect of combining attribution maps with the text signal. Second, we assess in-domain detection performance, where the proxy model matches the generator of the evaluated text. Third, we consider a unified setting in which samples from multiple LLMs are mixed during training, and generalization includes a held-out generator. Fourth, we evaluate robustness under adversarial attacks. Fifth, we examine cross-dataset generalization on an external benchmark.

\paragraph{Do attribution maps provide useful detection signals in encoder-decoder models?}

In encoder–decoder architectures, cross-attention models the dependencies between source and target tokens by enabling the decoder to focus selectively on the most relevant encoder states during generation.
In our task, we use the entire 128×128 attention matrix as the signal to be processed (as we selected samples of at most 128 tokens).
Table~\ref{tab:marian_all} reports results for AI-translation detection across three Marian MT-language pairs.
Using attribution maps alone (CNN), our method consistently outperforms the text-only baseline for all metrics, with gains of +3.6 F1 for ar-en (74.6 vs.\ 71.0), +6.7 for de-en (81.5 vs.\ 74.8), and +8.3 for fr-en (85.1 vs.\ 76.8).
Adding target-text features (CNN+text) yields a further, but smaller, improvement over CNN in every case (+2.0, +1.7, and +1.4 F1, respectively), suggesting that most discriminative signal is already captured by the attribution structure, while text provides complementary information.

\begin{table}[htbp]
\centering
\caption{Performance on Marian-MT generated translation and attribution maps. CNN and CNN+text configurations are based on \textsc{AEyeDE}.}
\label{tab:marian_all}
\resizebox{\textwidth}{!}{%
\begin{tabular}{lccccc|ccccc|ccccc}
\toprule
& \multicolumn{5}{c|}{\textbf{ar-en}}
& \multicolumn{5}{c|}{\textbf{de-en}}
& \multicolumn{5}{c}{\textbf{fr-en}} \\
\cmidrule(lr){2-6}\cmidrule(lr){7-11}\cmidrule(lr){12-16}
\textbf{Config}
& Acc & Prec & Rec & F1 & AUC
& Acc & Prec & Rec & F1 & AUC
& Acc & Prec & Rec & F1 & AUC \\
\midrule
AEyeDE(CNN)
& 68.9 & 63.0 & 91.4 & 74.6 & 77.7
& 79.0 & 72.6 & 93.0 & 81.5 & 87.3
& 83.6 & 78.0 & 93.6 & 85.1 & 91.0 \\
AEyeDE(CNN+text)
& \textbf{71.9} & \textbf{65.7} & \textbf{91.9} & \textbf{76.6} & \textbf{81.5}
& \textbf{81.1} & \textbf{74.9} & \textbf{93.6} & \textbf{83.2} & \textbf{89.2}
& \textbf{85.4} & \textbf{80.3} & \textbf{93.9} & \textbf{86.5} & \textbf{92.7} \\
\specialrule{0.2pt}{0.15em}{0.15em}
text (baseline)
& 61.9 & 57.3 & 93.4 & 71.0 & 71.5
& 69.5 & 63.7 & 90.5 & 74.8 & 77.5
& 72.4 & 66.2 & 91.4 & 76.8 & 81.4 \\
\bottomrule
\end{tabular}%
}
\end{table}

\begin{table}[htbp]
\centering
\caption{Results on \textit{individual} arrangement where each human and AI-generated attention attributions were extracted from the same model specified in the dataset to train the \textsc{AEyeDE} model.}
\resizebox{\textwidth}{!}{%
\begin{tabular}{lcccccc | cccccc}
\toprule
 & \multicolumn{6}{c|}{Cohere} & \multicolumn{6}{c}{GPT-neo} \\
 & Acc & Prec & Rec & F1 & AUC & TPR-FPR & Acc & Prec & Rec & F1 & AUC & TPR-FPR \\
\midrule
\textsc{AEyeDE} & \textbf{97.23} & \textbf{97.48} & 98.43 & \textbf{97.95} & \textbf{99.51} & \textbf{95.22} & \textbf{97.22} & 97.20 & 98.73 & \textbf{97.96} & \textbf{99.71} & \textbf{95.89} \\
Binoculars & 67.94 & 67.91 & 99.48 & 80.72 & 78.10 & 34.83 & 67.61 & 67.63 & \textbf{99.78} & 80.62 & 25.84 & 0.45 \\
Curvature & 83.27 & 86.29 & 89.39 & 87.81 & 90.86 & 47.01 & 78.91 & \textbf{98.22} & 70.03 & 81.76 & 79.70 & 65.47 \\
GLTR & 67.39 & 67.42 & \textbf{99.93} & 80.52 & 76.68 & 10.99 & 67.96 & 67.92 & 99.55 & 80.75 & 71.54 & 44.99 \\
LogRank & 67.39 & 67.42 & \textbf{99.93} & 80.52 & 75.49 & 10.24 & 69.58 & 69.31 & 98.58 & 81.39 & 73.95 & 36.70 \\
Rank & 67.39 & 67.42 & \textbf{99.93} & 80.52 & 53.66 & 0.00 & 73.11 & 77.14 & 85.50 & 81.11 & 79.05 & 21.52 \\
RoBERTa & 83.27 & 84.26 & 92.45 & 88.17 & 90.42 & 23.92 & 75.13 & 75.04 & 94.62 & 83.70 & 79.58 & 20.03 \\
\midrule
\midrule
 & \multicolumn{6}{c|}{Llama} & \multicolumn{6}{c}{Mistral} \\
 & Acc & Prec & Rec & F1 & AUC & TPR-FPR & Acc & Prec & Rec & F1 & AUC & TPR-FPR \\
\midrule
\textsc{AEyeDE} & \textbf{98.99} & \textbf{99.18} & 99.33 & \textbf{99.25} & \textbf{99.86} & \textbf{99.33} & \textbf{95.09} & \textbf{97.40} & 95.29 & \textbf{96.34} & \textbf{98.81} & \textbf{88.94} \\
Binoculars & 98.08 & 98.80 & 98.36 & 98.58 & 99.66 & 97.91 & 67.48 & 67.63 & 99.63 & 80.57 & 55.49 & 11.81 \\
Curvature & 67.37 & 67.44 & 99.85 & 80.51 & 78.67 & 39.69 & 67.68 & 67.68 & \textbf{100.00} & 80.72 & 70.32 & 49.10 \\
GLTR & 87.39 & 90.30 & 91.11 & 90.70 & 93.30 & 41.18 & 67.68 & 67.68 & \textbf{100.00} & 80.72 & 69.29 & 36.77 \\
LogRank & 85.93 & 90.08 & 88.94 & 89.51 & 92.16 & 38.04 & 67.93 & 68.00 & 99.40 & 80.75 & 69.01 & 28.03 \\
Rank & 67.57 & 67.54 & \textbf{100.00} & 80.63 & 51.41 & 0.00 & 67.83 & 67.80 & 99.93 & 80.79 & 59.71 & 0.00 \\
RoBERTa & 95.97 & 97.94 & 96.04 & 96.98 & 98.35 & 83.86 & 72.89 & 72.76 & 95.81 & 82.71 & 79.17 & 11.36 \\
\bottomrule
\end{tabular}
}
\label{tab:results_RAID_individual}
\end{table}

\begin{table}[htbp]
\centering
\caption{Results on the \textit{unified} arrangement where the samples corresponding to the Mistral model were excluded from the training of \textsc{AEyeDE}.}
\resizebox{\textwidth}{!}{%
\begin{tabular}{lcccccc | cccccc}
\toprule
 & \multicolumn{6}{c|}{Cohere} & \multicolumn{6}{c}{GPT-neo} \\
 & Acc & Prec & Rec & F1 & AUC & TPR-FPR & Acc & Prec & Rec & F1 & AUC & TPR-FPR \\
\midrule
\textsc{AEyeDE} & \textbf{72.38} & 90.98 & 65.55 & 76.19 & \textbf{80.67} & \textbf{58.67} & \textbf{72.25} & \textbf{96.57} & 61.06 & 74.82 & \textbf{78.86} & \textbf{58.97} \\
Binoculars & 67.29 & 67.57 & 99.03 & 80.33 & 74.21 & 49.33 & 67.46 & 67.51 & 99.85 & 80.55 & 74.43 & 43.57 \\
Curvature & 57.41 & 79.03 & 50.15 & 61.36 & 64.00 & 20.18 & 70.13 & 93.37 & 60.01 & 73.07 & 77.82 & 52.84 \\
GLTR & 67.44 & 67.44 & \textbf{100.00} & \textbf{80.55} & 76.32 & 26.53 & 67.51 & 68.11 & 97.53 & 80.21 & 73.42 & 15.10 \\
LogRank & 67.44 & 67.44 & \textbf{100.00} & \textbf{80.55} & 75.14 & 24.51 & 68.06 & 69.44 & 94.10 & 79.91 & 73.20 & 16.52 \\
Rank & 67.44 & 67.44 & \textbf{100.00} & \textbf{80.55} & 47.76 & 0.52 & 67.51 & 67.51 & \textbf{100.00} & \textbf{80.60} & 64.02 & 4.11 \\
RoBERTa & 69.66 & 80.87 & 72.05 & 76.21 & 77.34 & 35.50 & 69.48 & 75.83 & 80.42 & 78.06 & 77.17 & 38.27 \\
\midrule
\midrule
 & \multicolumn{6}{c|}{Llama} & \multicolumn{6}{c}{Mistral} \\
 & Acc & Prec & Rec & F1 & AUC & TPR-FPR & Acc & Prec & Rec & F1 & AUC & TPR-FPR \\
\midrule
\textsc{AEyeDE} & \textbf{71.96} & \textbf{92.59} & 63.53 & 75.35 & \textbf{79.29} & \textbf{56.80} & \textbf{71.83} & \textbf{95.14} & 61.51 & 74.72 & \textbf{78.86} & \textbf{58.59} \\
Binoculars & 67.52 & 67.61 & 99.55 & 80.53 & 75.99 & 50.45 & 67.48 & 67.70 & 99.33 & 80.52 & 72.59 & 46.49 \\
Curvature & 63.04 & 82.42 & 57.47 & 67.72 & 69.42 & 27.50 & 59.28 & 81.84 & 51.20 & 62.99 & 65.14 & 28.48 \\
GLTR & 67.47 & 67.47 & \textbf{100.00} & \textbf{80.58} & 74.20 & 22.12 & 67.63 & 67.66 & 99.93 & 80.69 & 75.19 & 24.07 \\
LogRank & 67.47 & 67.47 & \textbf{100.00} & \textbf{80.58} & 72.75 & 18.91 & 67.68 & 67.68 & \textbf{100.00} & \textbf{80.72} & 74.63 & 20.55 \\
Rank & 67.47 & 67.47 & \textbf{100.00} & \textbf{80.58} & 45.44 & 0.52 & 67.68 & 67.68 & \textbf{100.00} & \textbf{80.72} & 54.88 & 0.22 \\
RoBERTa & 70.20 & 77.48 & 78.70 & 78.09 & 77.92 & 39.76 & 71.12 & 79.89 & 76.61 & 78.21 & 78.49 & 38.86 \\
\bottomrule
\end{tabular}
}
\label{tab:results_RAID_unified}
\end{table}

\begin{table}[htbp]
\centering
\caption{Results on HC3 using GPT-neo to extract attention attributions.}
\small
\begin{tabular}{lcccccc}
\toprule
 & \multicolumn{6}{c}{GPT-neo} \\
 & Acc & Prec & Rec & F1 & AUC & TPR-FPR \\
\midrule
\textsc{AEyeDE}    & 96.85 & 95.68 & 98.27 & 96.96 & 99.51 & 93.50 \\
Curvature & \textbf{98.40} & \textbf{99.83} & 97.04 & \textbf{98.42} & 99.22 & 97.45 \\
GLTR      & 98.25 & 98.89 & 97.60 & 98.24 & 99.17 & 97.60 \\
LogRank   & 97.95 & 99.69 & 96.20 & 97.91 & 99.03 & 97.50 \\
Rank      & 83.40 & 81.63 & 86.20 & 83.85 & 91.22 & 36.60 \\
RoBERTa   & 97.73 & 96.70 & \textbf{98.93} & 97.80 & \textbf{99.57} & \textbf{97.78} \\
\bottomrule
\end{tabular}
\label{tab:results-HC3}
\end{table}

\paragraph{How effective is \textsc{AEyeDE} when the proxy model matches the generator family?}

Using the RAID dataset, we train a separate detector for each generator family using only samples generated by that family, along with the corresponding matched human texts (e.g., we train Llama samples with the attributions extracted from the Llama model). We refer to this configuration as the \textit{individual} setting. For the model-specific baselines (Curvature, GLTR, Rank, and LogRank), the underlying scores are also extracted from the corresponding generator model. It should be noted that both LLMs used for binoculars are from Llama family.

In the \textit{individual} setting (Table~\ref{tab:results_RAID_individual}), \textsc{AEyeDE} achieves near-ceiling performance across all four generator families, with F1 ranging from 96.34 to 99.25 and AUC from 98.81 to 99.86, and it is the top-performing method on every generator. The gains are especially large for GPT-neo and Mistral. For GPT-neo, the strongest baseline reaches 83.70 F1 (RoBERTa), whereas \textsc{AEyeDE} attains 97.96. For Mistral, the strongest baseline reaches 82.71 F1 (RoBERTa), compared with 96.34 for \textsc{AEyeDE}. On Cohere, the strongest baselines are more competitive, with RoBERTa reaching 88.17 F1 and Curvature 87.81, but \textsc{AEyeDE} still improves substantially to 97.95. The baseline behavior is more heterogeneous for Llama. Several methods perform strongly in this case, including Binoculars (98.58 F1, 99.66 AUC), RoBERTa (96.98 F1), and GLTR (90.70 F1). Even under this more favorable setting for the baselines, \textsc{AEyeDE} remains best overall, achieving 99.25 F1, 99.86 AUC, and 99.33 on the low-FPR operating-point metric. 

Furthermore, a broader pattern is that likelihood- and rank-based baselines, especially on Cohere, GPT-neo, and Mistral, yield recall near 100\% but precision close to the class prior (\(\sim 67\%\)). In contrast, \textsc{AEyeDE} maintains both high precision and high recall across all generators, which suggests that the attention-based attribution signals are highly discriminative when the proxy model matches the generator family.

Likewise, we train and test AEyeDE on the HC3 dataset. Table~\ref{tab:results-HC3} reports results on the \textsc{HC3} dataset using attribution maps extracted from GPT-neo as the proxy model, as it is generated only by the GPT family. Overall, \textsc{HC3} appears substantially less challenging than \textsc{RAID}, as nearly all methods achieve very high performance. \textsc{AEyeDE} remains competitive, achieving 96.85 accuracy, 95.68 precision, 98.27 recall, 96.96 F1, and 99.51 AUC. Although Curvature obtains the best F1 and RoBERTa achieves the highest AUC, \textsc{AEyeDE} is notable because it reaches near-ceiling performance using a fundamentally different signal. Moreover, the comparison with RoBERTa should be interpreted cautiously, since HC3 was part of that model’s training data. We employ the trained models on the RAID dataset along with the model trained on HC3 data, to analyze informative motif patterns for human vs. AI-generated text (See \ref{sec-disccusion}).

\paragraph{Do attribution-based representations transfer to unseen generators?} In a different setup, we evaluate how well our proposed model can generalize to an unseen generator at test time. We train on a \textit{unified} mixture of all RAID generators except Mistral, but the test set includes Mistral's generation. To control for training set size, we subsample each included generator's generated text to match the per-model training budget used in the \textit{individual} setting. This splitting strategy assesses whether representations learned from a subset of generator families transfer to an unseen model at test time.

In the \textit{unified} setting (Table~\ref{tab:results_RAID_unified}), performance is lower than in the generator-matched \textit{individual} setting, reflecting the increased heterogeneity of pooled training and the greater difficulty of cross-generator transfer. Even so, \textsc{AEyeDE} achieves the best accuracy, AUC, and TPR@FPR=0.01 for every generator family. Its AUC ranges from 78.86 to 80.67, and its TPR@FPR=0.01 ranges from 56.80 to 58.97, consistently exceeding all baselines.
At the same time, \textsc{AEyeDE} adopts a markedly more conservative operating point than most competing methods. Across all generators, it maintains very high precision (90.98-96.57) but substantially lower recall (61.06-65.55), which leads to F1 scores in the 74.72-76.19 range. In contrast, Binoculars, GLTR, LogRank, and Rank operate in an almost-all-positive regime, with recall near 100\% but precision close to the class prior (\(\sim67\%\)). Their F1 values therefore appear superficially strong (\(\sim80.2\)-80.7), but this comes at the cost of many false positives. RoBERTa is the strongest balanced baseline, with F1 between 76.21 and 78.21 and AUC between 77.17 and 78.49, yet it still trails \textsc{AEyeDE} on the threshold-independent metrics and on the low-FPR operating point.

These results suggest that attribution-based representations do transfer across generator families, but they do so conservatively: when trained on mixed generators, \textsc{AEyeDE} becomes more selective, flagging fewer human texts incorrectly while missing a larger fraction of AI-generated samples at the default threshold. Importantly, its consistent advantage in AUC and TPR@FPR=0.01 indicates that it provides the best overall separation between human and machine text under cross-generator transfer.

\paragraph{How robust is AEyeDE to adversarial perturbations?} AI-generated text detectors are known to be brittle under adversarial perturbations~\citep{NEURIPS2023_575c4500}. We evaluate robustness to the paraphrasing and alternative spelling attacks provided in the RAID dataset for both the individual and unified settings.\footnote{Due to computational constraints, we report these adversarial experiments for GPT-neo only.}

In the individual setting (Table~\ref{tab:results-RAID-pph-ind}), \textsc{AEyeDE} achieves the highest F1 (81.50) and AUC (77.64), with a substantial precision advantage over all baselines (74.78 vs.\ 68.87 for the next-best method, RoBERTa)\footnote{The paraphrases are generated by a T5 model, so the adversarial samples remain machine-generated, although they are out-of-distribution relative to the training data. Importantly, adversarial samples are used only at test time, meaning that \textsc{AEyeDE} is not trained on adversarial examples.}. Binoculars, GLTR, LogRank, and Rank exhibit near-perfect recall ($\geq 99.48\%$) but precision close to the class prior (\textasciitilde67\%), indicating a near-all-positive prediction behaviour. Curvature attains the highest accuracy (76.13), but its recall drops sharply (54.13), yielding by far the lowest F1 (55.69). In contrast, \textsc{AEyeDE} maintains the best precision-recall tradeoff (74.78 / 89.54), suggesting that attribution-based structure appears more robust to paraphrastic surface perturbations in the individual setting.

In the unified adversarial setting (Table~\ref{tab:results-RAID-pph-unified}), the distinction between methods becomes much smaller in terms of thresholded metrics: accuracy remains near the majority-class baseline (\textasciitilde67\%) and F1 values cluster tightly (78.92-80.45). \textsc{AEyeDE} still attains the highest F1 (80.45), but only by a narrow margin, and its precision/recall profile (67.50 / 99.55) indicates that it too shifts toward the near-all-positive trend. Moreover, its AUC drops to 62.21, below RoBERTa (72.87) and Curvature (70.38). Overall, these results show that paraphrasing remains a difficult adversarial condition for all detectors. \textsc{AEyeDE} retains a clear advantage in the individual setting, but that advantage disappears in the unified setting.

We also evaluate robustness to alternative-spelling perturbations, motivated by the fact that spelling errors are relatively uncommon in clean LLM outputs. In the RAID dataset, this attack is constructed by randomly modifying characters in the original samples. In the individual setting (Table~\ref{tab:results-individual-spelling}), \textsc{AEyeDE} remains highly effective under this attack, achieving the best F1 (98.47), AUC (99.75), and TPR@FPR=0.01 (96.26). It outperforms all baselines by wide margins, with the next-best F1 being 84.06 for RoBERTa. Several baselines, especially Binoculars and GLTR, operate close to an all-positive score, with near-perfect recall but substantially weaker precision and poor low-FPR detection. These results suggest that character-level spelling perturbations affect surface-level statistics more than the attribution patterns used by \textsc{AEyeDE}.

In the unified setting (Table~\ref{tab:results-unified-spelling}), performance decreases across methods, but \textsc{AEyeDE} still achieves the best accuracy (70.34), F1 (81.33), AUC (83.48), and TPR@FPR=0.01 (59.10). Its low-FPR true-positive rate exceeds that of the next-best method, Curvature (48.87), by more than 10 points. Curvature attains the highest precision (93.88), but its recall drops to 56.47, yielding the lowest F1 among the reported methods (70.52). Overall, these results suggest that attribution-based detection particularly works well against this type of character-level perturbation. One possible explanation is that subword tokenization does not fully obscure the token-level relationships captured by the attribution maps.

\begin{table}[htbp]
\centering
\scriptsize
\setlength{\tabcolsep}{3pt}
\begin{minipage}{0.48\textwidth}
\centering
\caption{Results under the RAID paraphrasing adversarial in the \textit{individual} setting with GPT-neo.}
\begin{tabular}{lcccccc}
\toprule
\textbf{Method} & \multicolumn{6}{c}{\textbf{GPT-neo}} \\
 & Acc & Prec & Rec & F1 & AUC & TPR \\
\midrule
\textsc{AEyeDE} & 72.68 & \textbf{74.78} & 89.54 & \textbf{81.50} & \textbf{77.64} & 23.17 \\
Binoculars & 67.20 & 67.20 & \textbf{100.00} & 80.38 & 15.15 & 0.00 \\
Curvature & \textbf{76.13} & 57.34 & 54.13 & 55.69 & 72.38 & 26.67 \\
GLTR & 67.05 & 67.17 & 99.70 & 80.26 & 65.95 & 26.98 \\
LogRank & 67.10 & 67.19 & 99.78 & 80.30 & 66.32 & \textbf{28.48} \\
Rank & 66.90 & 67.12 & 99.48 & 80.16 & 68.99 & 13.30 \\
RoBERTa & 68.96 & 68.87 & 98.21 & 80.96 & 68.73 & 5.08 \\
\bottomrule
\end{tabular}
\label{tab:results-RAID-pph-ind}
\end{minipage}
\hfill
\begin{minipage}{0.48\textwidth}
\centering
\caption{Results under the RAID paraphrasing adversarial in the \textit{unified} setting with GPT-neo.}
\begin{tabular}{lcccccc}
\toprule
\textbf{Method} & \multicolumn{6}{c}{\textbf{GPT-neo}} \\
 & Acc & Prec & Rec & F1 & AUC & TPR \\
\midrule
\textsc{AEyeDE} & \textbf{67.52} & 67.50 & 99.55 & \textbf{80.45} & 62.21 & 4.80 \\
Binoculars & 67.12 & 67.12 & \textbf{100.00} & 80.33 & 66.26 & 18.98 \\
Curvature & 67.02 & 67.12 & 99.70 & 80.23 & 70.38 & \textbf{19.88} \\
GLTR & 67.17 & 67.15 & \textbf{100.00} & 80.35 & 66.77 & 6.98 \\
LogRank & 67.17 & 67.15 & \textbf{100.00} & 80.35 & 65.50 & 10.65 \\
Rank & 67.12 & 67.12 & \textbf{100.00} & 80.33 & 57.02 & 2.55 \\
RoBERTa & 66.21 & \textbf{67.89} & 94.22 & 78.92 & \textbf{72.87} & 16.58 \\
\bottomrule
\end{tabular}
\label{tab:results-RAID-pph-unified}
\end{minipage}
\end{table}

\begin{table}[htbp]
\centering
\scriptsize
\setlength{\tabcolsep}{3pt}
\begin{minipage}{0.48\textwidth}
\centering
\caption{Results under the RAID alternative-spelling adversarial in the \textit{individual} setting with GPT-neo.}
\begin{tabular}{lcccccc}
\toprule
\textbf{Method} & \multicolumn{6}{c}{\textbf{GPT-neo}} \\
 & Acc & Prec & Rec & F1 & AUC & TPR \\
\midrule
\textsc{AEyeDE} & \textbf{97.93} & \textbf{98.07} & 98.88 & \textbf{98.47} & \textbf{99.75} & \textbf{96.26} \\
Binoculars & 67.44 & 67.44 & \textbf{100.00} & 80.55 & 33.56 & 2.02 \\
Curvature & 77.37 & 96.25 & 69.13 & 80.47 & 77.02 & 64.65 \\
GLTR & 67.39 & 67.42 & 99.93 & 80.52 & 69.39 & 42.38 \\
LogRank & 68.65 & 68.92 & 97.46 & 80.74 & 72.28 & 32.36 \\
Rank & 74.55 & 77.93 & 86.85 & 82.15 & 78.11 & 15.47 \\
RoBERTa & 76.21 & 76.66 & 93.05 & 84.06 & 80.56 & 6.28 \\
\bottomrule
\end{tabular}
\label{tab:results-individual-spelling}
\end{minipage}
\hfill
\begin{minipage}{0.5\textwidth}
\centering
\caption{Results under the RAID alternative-spelling adversarial in the \textit{unified} setting with GPT-neo.}
\begin{tabular}{lcccccc}
\toprule
\textbf{Method} & \multicolumn{6}{c}{\textbf{GPT-neo}} \\
 & Acc & Prec & Rec & F1 & AUC & TPR \\
\midrule
\textsc{AEyeDE} & \textbf{70.34} & 70.58 & 95.94 & \textbf{81.33} & \textbf{83.48} & \textbf{59.10} \\
Binoculars & 67.30 & 67.32 & 99.92 & 80.44 & 70.60 & 37.59 \\
Curvature & 68.22 & \textbf{93.88} & 56.47 & 70.52 & 75.00 & 48.87 \\
GLTR & 64.78 & 78.98 & 64.96 & 71.29 & 69.55 & 16.17 \\
LogRank & 65.18 & 80.00 & 64.36 & 71.33 & 69.77 & 14.51 \\
Rank & 67.31 & 67.31 & \textbf{100.00} & 80.46 & 67.18 & 4.21 \\
RoBERTa & 69.74 & 78.33 & 76.09 & 77.19 & 76.00 & 8.05 \\
\bottomrule
\end{tabular}
\label{tab:results-unified-spelling}
\end{minipage}
\end{table}

\paragraph{Does AEyeDE generalize to a fully external dataset?} To complement the adversarial evaluations and assess generalization on a fully external test set, we evaluate the methods on Beemo (Table~\ref{tab:results-beemo}), which is unseen during training, using Mistral-7B-Instruct as the proxy model~\footnote{In our experiments, Mistral scored best among the other generators for this task.}. \textsc{AEyeDE} achieves the highest accuracy (66.36), precision (62.16), F1 (71.32), and AUC (74.43), outperforming all other methods on these metrics. Several baselines exhibit thresholded behavior close to an all-positive classifier, with accuracy near 50-51\%, precision near the class prior (\textasciitilde50\%), and recall above 96\%, indicating poor calibration at the chosen operating point. RoBERTa performs somewhat better (56.03 accuracy, 68.66 F1) but still trails \textsc{AEyeDE} by 2.66 F1 points. Notably, GLTR achieves a nearly identical AUC (74.38) and the highest TPR@FPR=0.01 (15.46, compared with 14.55 for \textsc{AEyeDE}), although its accuracy and precision remain close to chance at the selected threshold. Overall, these results indicate that the attribution-based signal transfers well across datasets and provides the strongest overall cross-dataset performance, even though some baselines remain competitive on ranking-based metrics such as AUC or low-FPR TPR. It is also noteworthy that these results are obtained with a relatively small training set, suggesting that the proposed framework can extract useful discriminative structure even under limited-data conditions.

\begin{table}[htbp]
\centering
\caption{Cross-dataset evaluation on Beemo using Mistral as the proxy model.}
\small
\begin{tabular}{lcccccc}
\toprule
\textbf{Method} & \multicolumn{6}{c}{\textbf{Mistral}} \\
 & Acc & Prec & Rec & F1 & AUC & TPR \\
\midrule
\textsc{AEyeDE} & \textbf{66.36} & \textbf{62.16} & 83.64 & \textbf{71.32} & \textbf{74.43} & 14.55 \\
Binoculars & 51.36 & 50.71 & 96.36 & 66.45 & 63.16 & 6.36 \\
Curvature & 49.44 & 49.74 & 98.73 & 66.16 & 50.47 & 9.76 \\
GLTR & 51.40 & 50.74 & \textbf{98.88} & 67.07 & 74.38 & \textbf{15.46} \\
LogRank & 51.27 & 50.68 & 98.47 & 66.92 & 74.25 & 13.01 \\
Rank & 50.92 & 50.50 & 97.71 & 66.59 & 64.16 & 3.05 \\
RoBERTa & 56.03 & 53.37 & 96.24 & 68.66 & 69.17 & 1.02 \\
\bottomrule
\end{tabular}
\label{tab:results-beemo}
\end{table}

\section{Analysis of Patch-Level Motifs in Attribution Maps}
\label{sec-disccusion}
Building on the attribution encoder $E_{\text{attr}}$ (Sec.~\ref{sec:framework}), we investigated whether the detector exploits \emph{localized} and \emph{repeatable} visual motifs in attribution maps that are characteristic of human-written versus AI-generated text.
Recall that after the final convolutional stage, each block $A_k\in\mathbb{R}^{128\times128}$ is mapped to a feature map of spatial size $16\times16$ with $256$ channels.
We denote this last feature map by
\[
F_k \in \mathbb{R}^{256\times 16\times 16}.
\]
Because $128/16=8$, each feature map cell $(u,v)$ corresponds to an $8\times8$ patch of the original block $A_k$ (shown in Fig.~\ref{fig:architecture}).
Let $P_{k,u,v}\in\mathbb{R}^{8\times 8}$ be this patch, and define its representation, obtained after the last CNN convolutional stage, as:
\[
z_{k,u,v} \;=\; F_k[:,u,v] \in \mathbb{R}^{256}.
\]
We denote $\{z_{k,u,v}\}$ as an embedding space of patches produced by the detector model.

\begin{figure}[!htbp]
\centering

\begin{subfigure}[t]{0.32\textwidth}
  \centering
  \includegraphics[width=\linewidth]{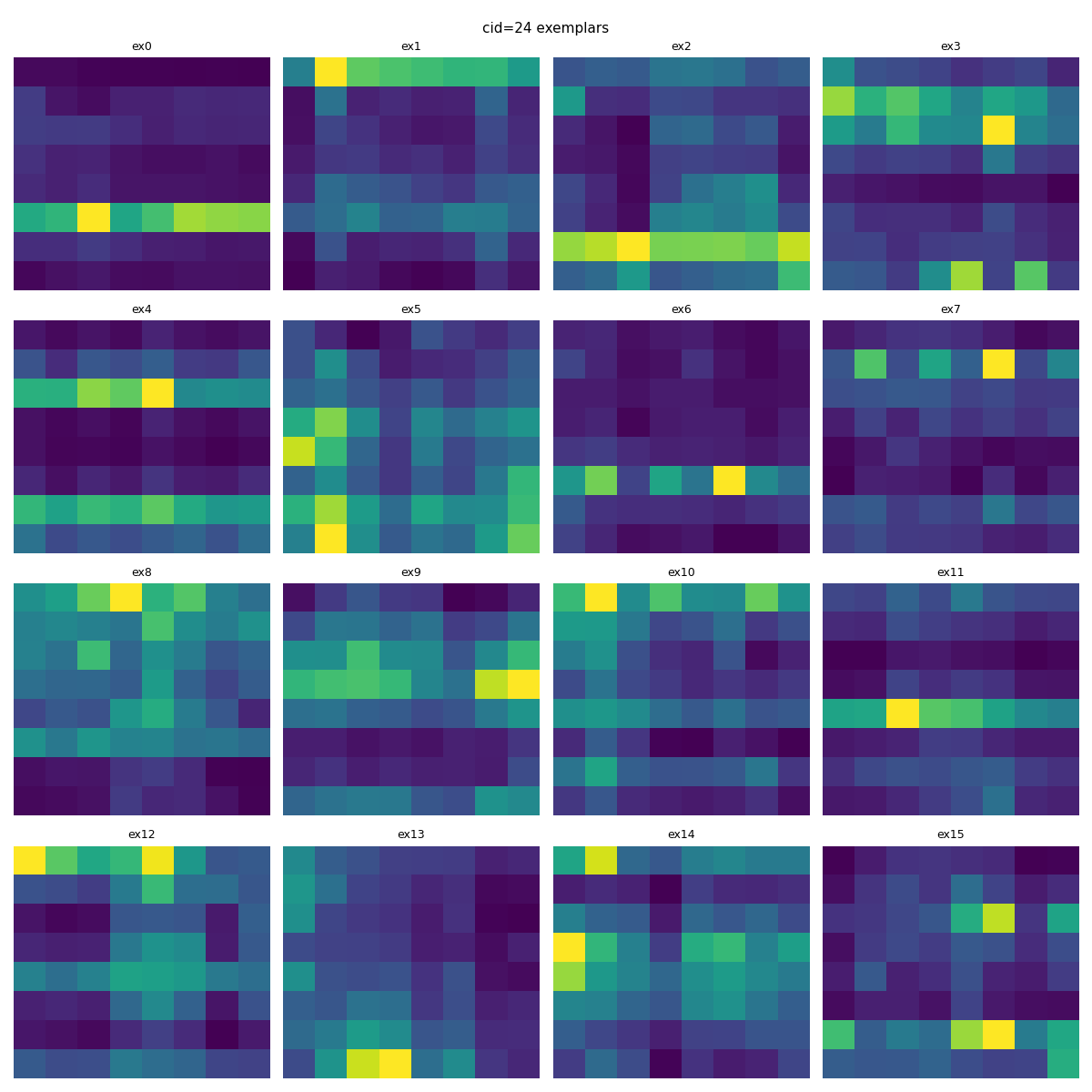}
  \caption{HC3 (individual), proxy: GPT-neo. Top cluster by $\Delta \bar r$ (machine-skewed).}
  \label{fig:motif-hc3-std-gpt-machine}
\end{subfigure}\hfill
\begin{subfigure}[t]{0.32\textwidth}
  \centering
  \includegraphics[width=\linewidth]{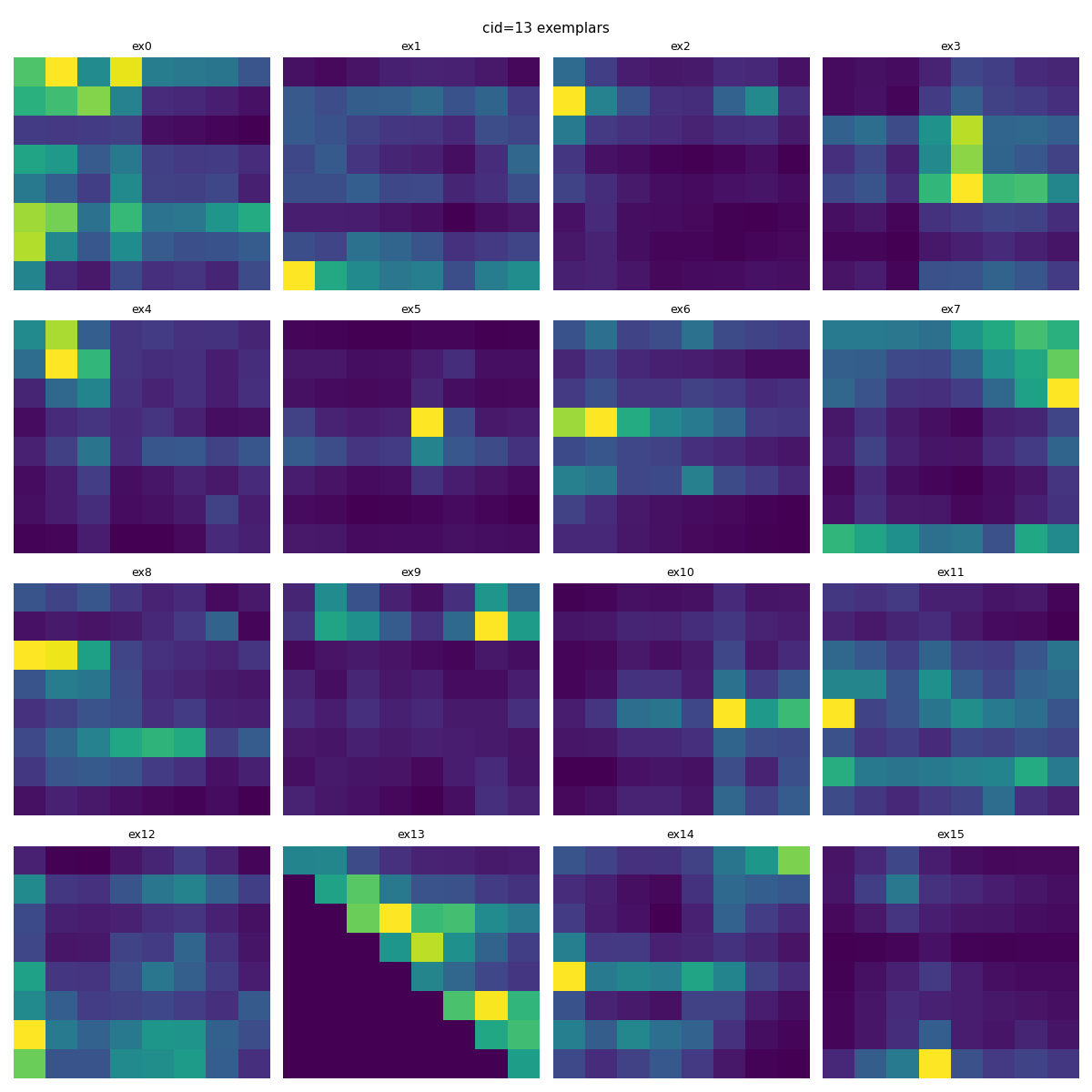}
  \caption{RAID (individual), proxy: Cohere. Top cluster by $\Delta \bar r$ (human-skewed).}
  \label{fig:motif-unified-cohere-human}
\end{subfigure}\hfill
\begin{subfigure}[t]{0.32\textwidth}
  \centering
  \includegraphics[width=\linewidth]{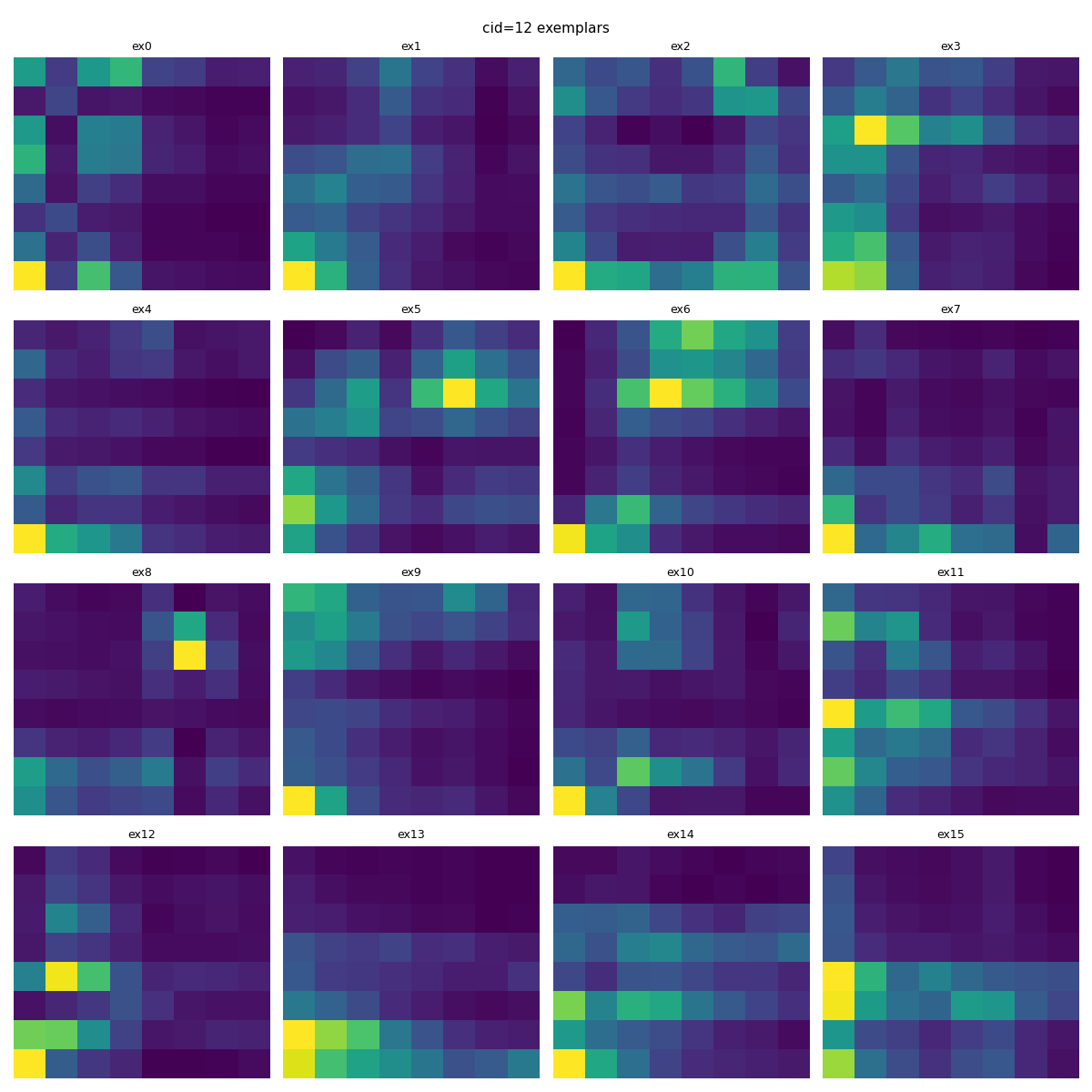}
  \caption{RAID (individual), proxy: Llama. Top cluster by $\Delta \bar r$ (human-skewed).}
  \label{fig:motif-std-llama-human}
\end{subfigure}

\vspace{2mm}

\begin{subfigure}[t]{0.32\textwidth}
  \centering
  \includegraphics[width=\linewidth]{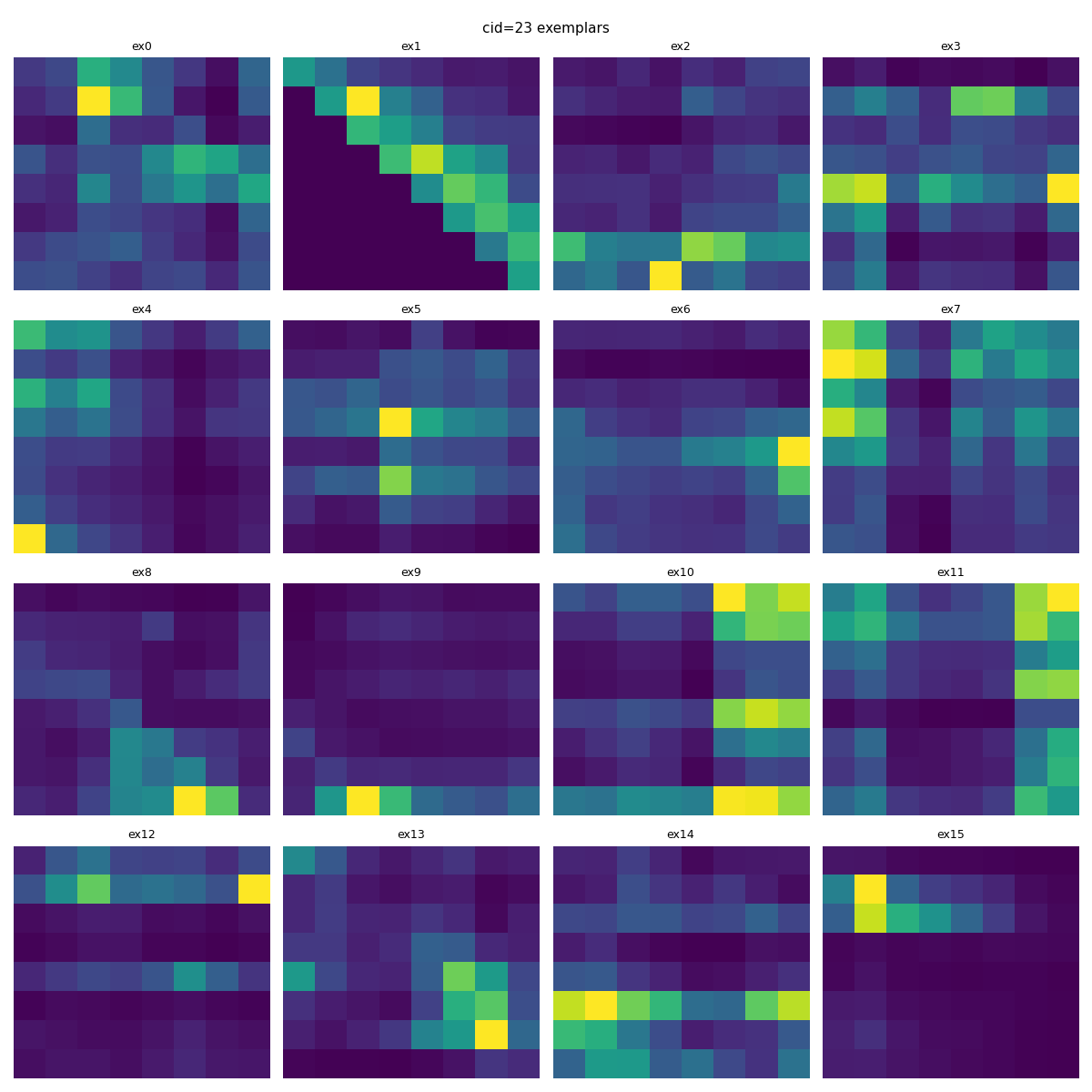}
  \caption{RAID (individual), proxy: Mistral. Top cluster by $\Delta \bar r$ (human-skewed).}
  \label{fig:motif-std-mistral-human}
\end{subfigure}\hfill
\begin{subfigure}[t]{0.32\textwidth}
  \centering
  \includegraphics[width=\linewidth]{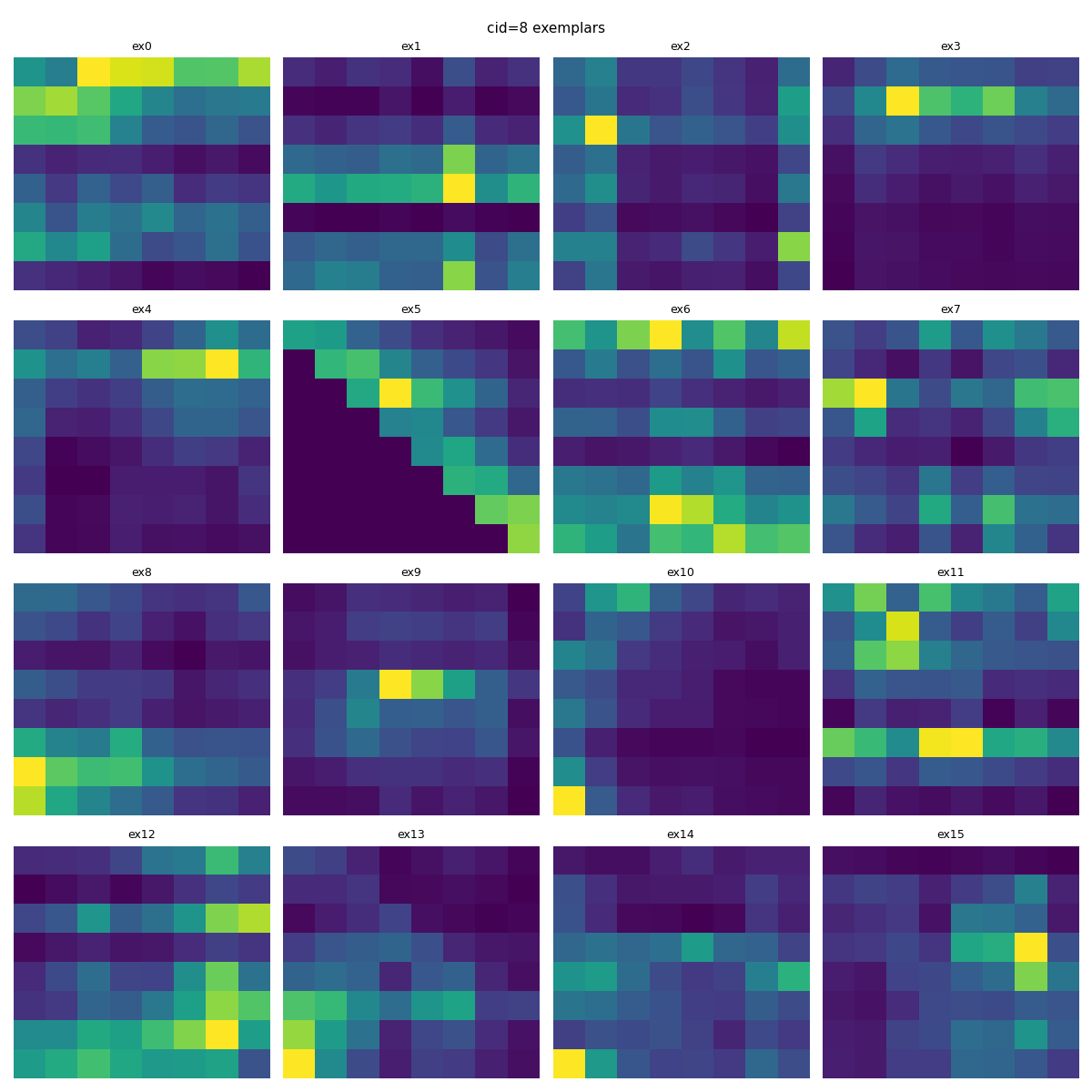}
  \caption{RAID (individual), proxy: Cohere. Top cluster by $\Delta \bar r$ (human-skewed).}
  \label{fig:motif-std-cohere-human}
\end{subfigure}\hfill
\begin{subfigure}[t]{0.32\textwidth}
  \centering
  \includegraphics[width=\linewidth]{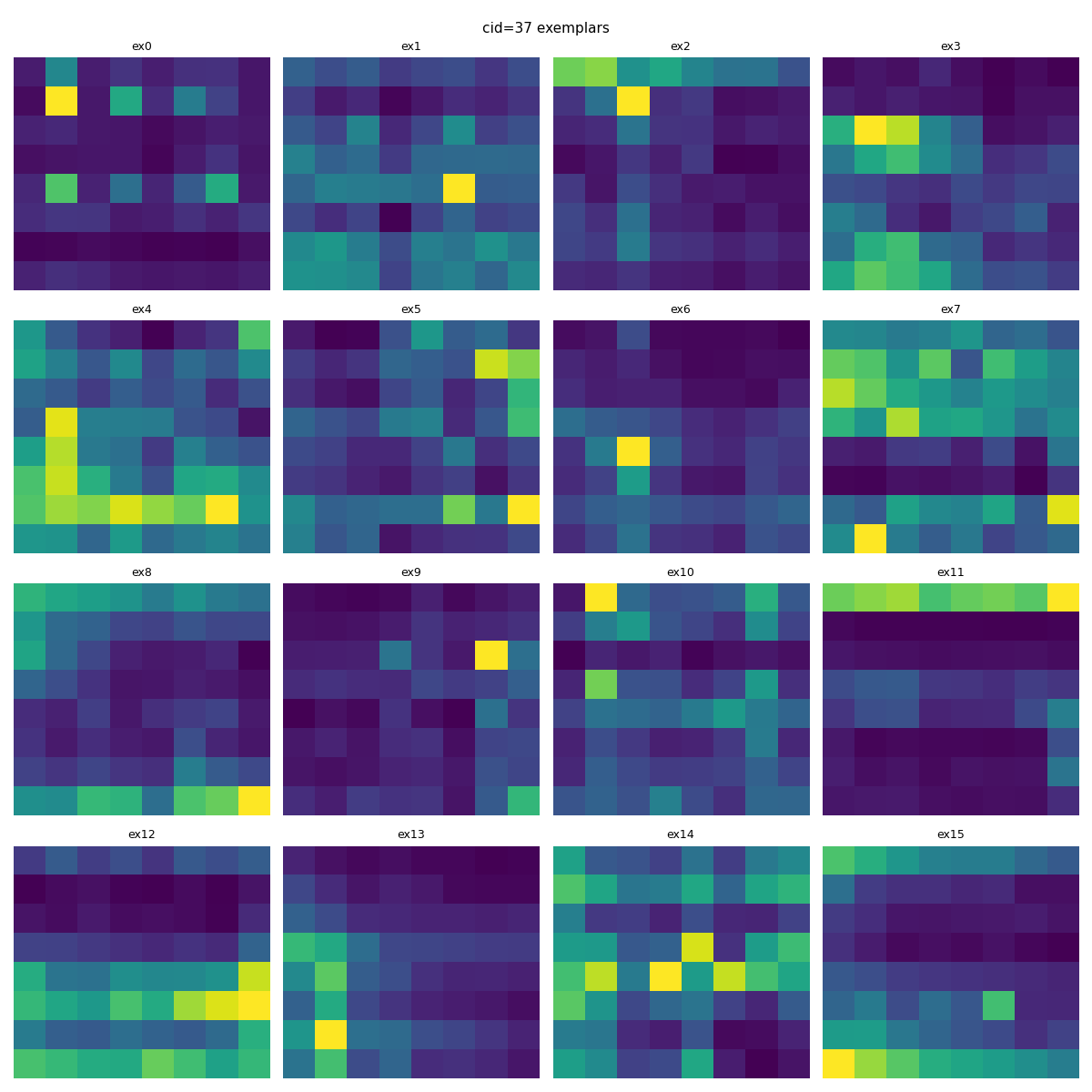}
  \caption{RAID (unified), proxy: GPT-neo. Top cluster by $\Delta \bar r$ (human-skewed).}
  \label{fig:motif-unified-gpt-human}
\end{subfigure}

\vspace{2mm}

\begin{subfigure}[t]{0.32\textwidth}
  \centering
  \includegraphics[width=\linewidth]{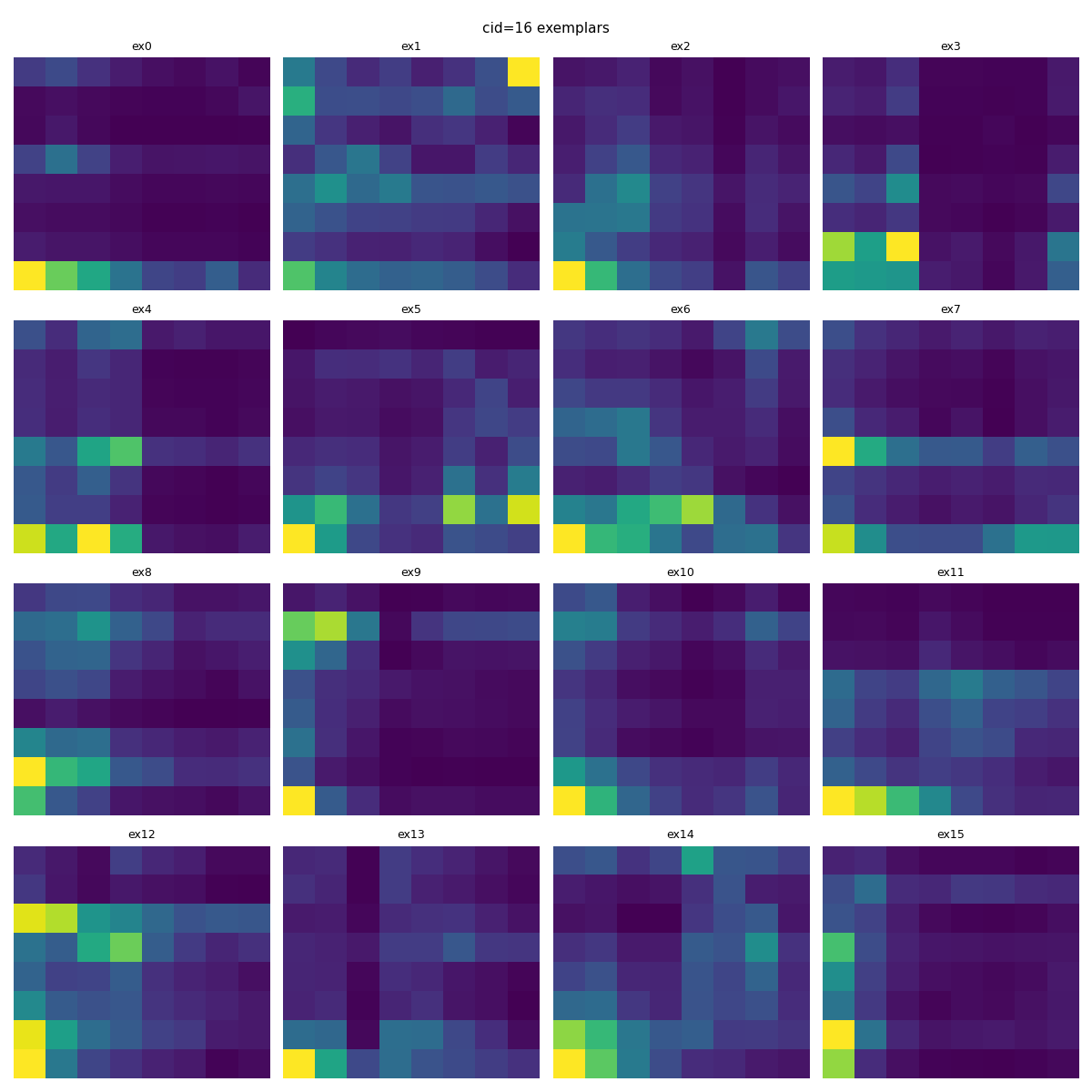}
  \caption{RAID (unified), proxy: Llama. Top cluster by $\Delta \bar r$ (human-skewed).}
  \label{fig:motif-unified-llama-human}
\end{subfigure}\hfill
\begin{subfigure}[t]{0.32\textwidth}
  \centering
  \includegraphics[width=\linewidth]{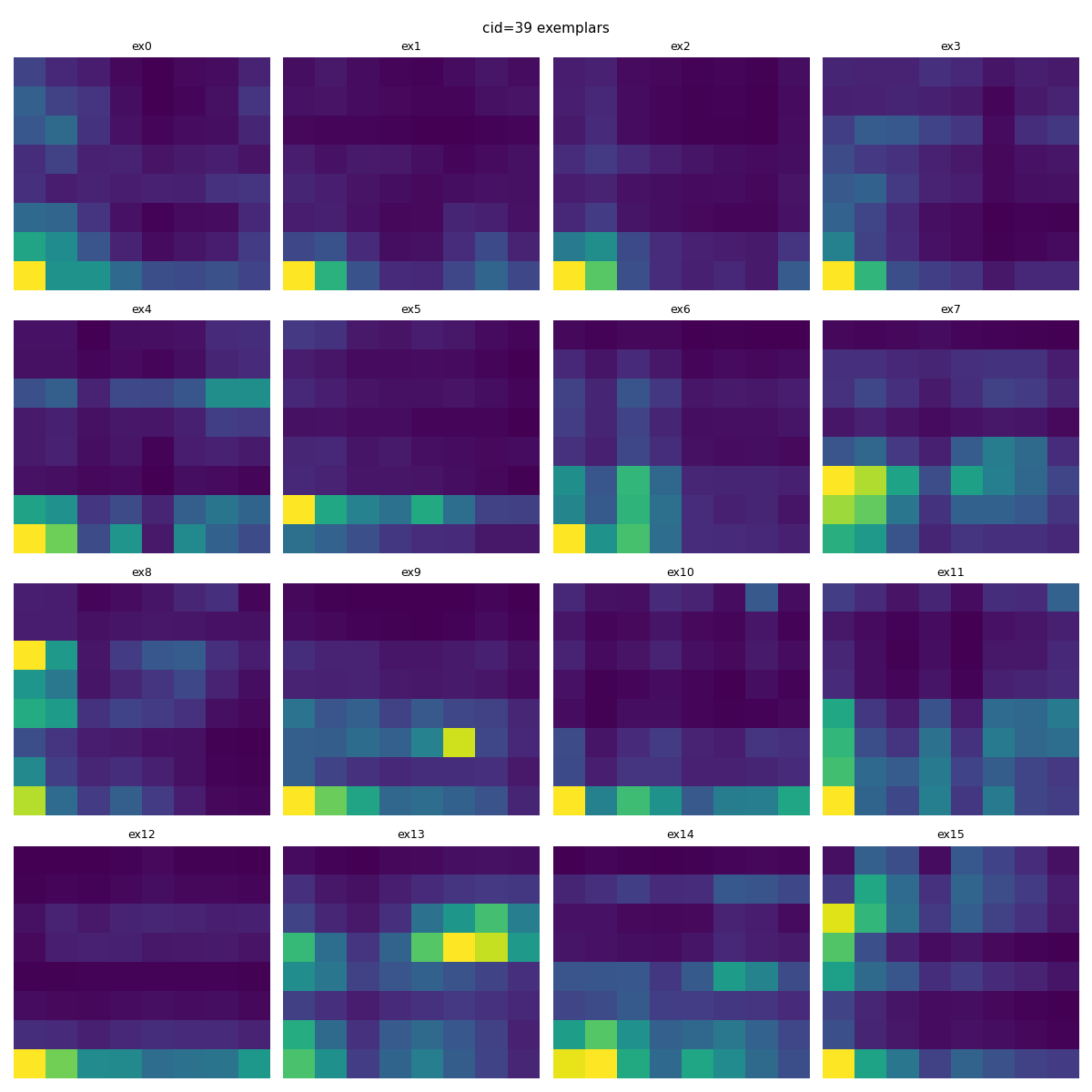}
  \caption{RAID (unified), proxy: Mistral. Top cluster by $\Delta \bar r$ (human-skewed).}
  \label{fig:motif-unified-mistral-human}
\end{subfigure}\hfill
\begin{subfigure}[t]{0.32\textwidth}
  \centering
  \includegraphics[width=\linewidth]{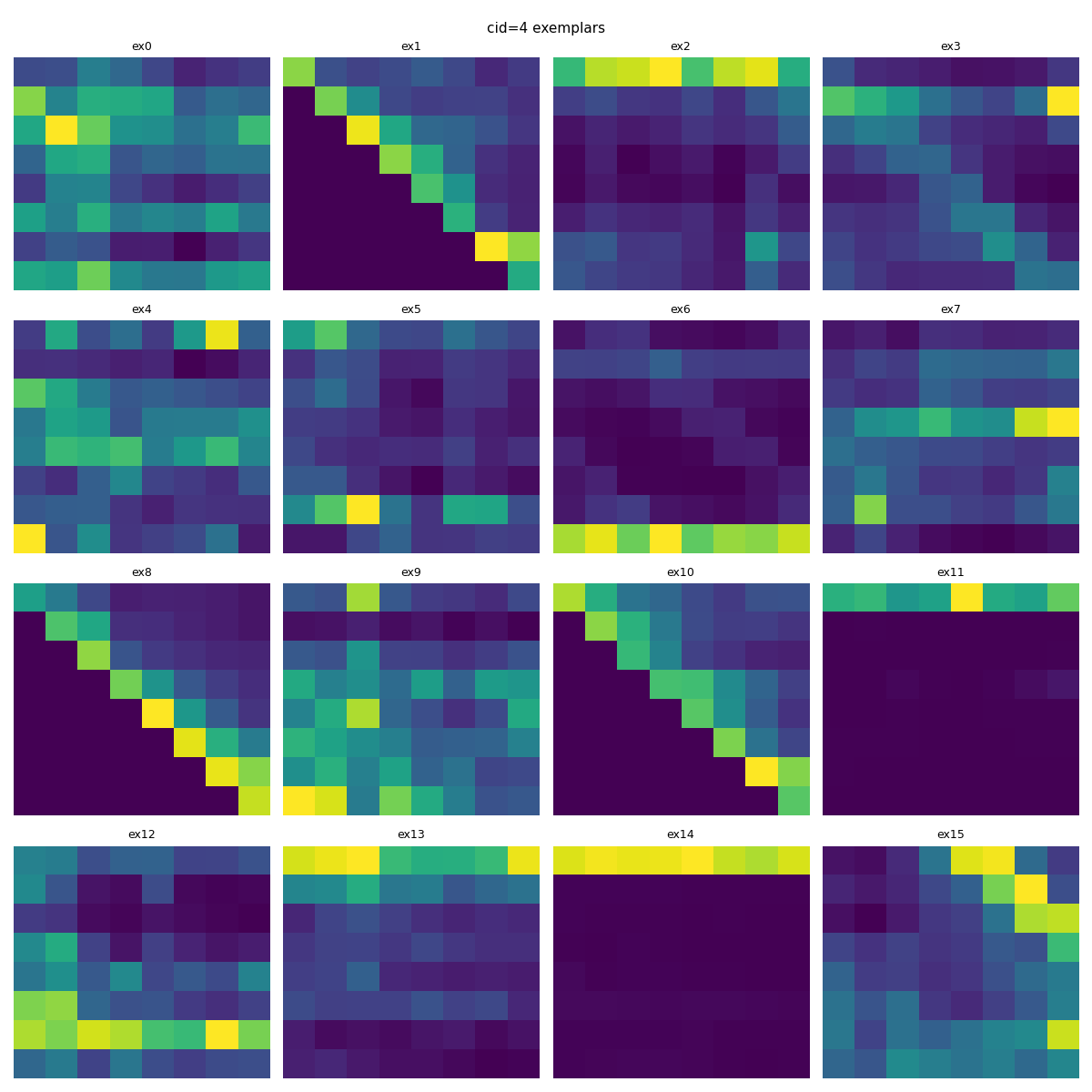}
  \caption{RAID (individual), proxy: GPT-neo. Top cluster by $\Delta \bar r$ (human-skewed).}
  \label{fig:motif-std-gpt-human}
\end{subfigure}

\caption{Examples of the top motif cluster (by absolute mean prevalence gap $\Delta \bar r$ between gold-machine and gold-human) for each dataset/proxy-model configuration. Each panel shows representative $8{\times}8$ patches (z-normalized) from the corresponding cluster.}
\label{fig:appendix-top-motif-clusters}
\end{figure}

\paragraph{Patch selection and clustering.}
To avoid padding-only and non-informative constant patches, we only keep patches that (i) correspond to non-padding entries under the mask $M=m_xm_y^\top$, and (ii) pass a minimal informativeness threshold based on mean and standard deviation inside the patch.
We keep patches with $|\mu_{k,u,v}|>\tau_\mu$(0.01) and $\sigma_{k,u,v}>\tau_\sigma$(0.001). Then, we  cluster their retained embeddings $\{z_{k,u,v}\}$ using HDBSCAN, producing assignments of each patch to the corresponding cluster
\begin{equation}
c_{k,u,v}\in\{1,\dots,C\}\cup\{-1\}.
\end{equation}
where $-1$ is unclustered noise.
Each cluster can be interpreted as a \emph{motif family}: a set of patches that the trained encoder maps to nearby representations.

\paragraph{Sample-normalized motif rates.}
A direct comparison of raw motif counts is complicated by two sources:
(i) datasets can be class-imbalanced (e.g., \textsc{RAID}), and (ii) the number of valid patches retained after preprocessing varies greatly across samples due to length and informativity selection.
To obtain a metric that is comparable across samples, for each sample $s$ we denote by $\mathcal{P}(s)$ the set of all retained patches extracted from that sample (across all diagonal blocks), and define the per-sample \emph{motif rate} for cluster $c$ as
\begin{equation}
r_s(c)
\;=\;
\frac{1}{|\mathcal{I}_s|}
\sum_{(k,u,v)\in \mathcal{I}_s}
\mathbb{I}\!\left[c_{k,u,v}=c\right],
\label{eq:per-sample-rate}
\end{equation}
Here $\mathcal{I}_s$ is the set of retained patch indices $(k,u,v)$ from sample $s$, and let $c_{k,u,v}$ be the patch-cluster assignment. That is, $r_s(c)$ is the fraction of all retained patches in sample $s$ that belong to cluster $c$.

For a group of samples $\mathcal{S}_g$ (defined below), we summarize prevalence by the mean rate
\begin{equation}
\bar r_g(c)\;=\;\frac{1}{|\mathcal{S}_g|}\sum\nolimits_{s\in\mathcal{S}_g} r_s(c).
\end{equation}
In our analysis, we report $\bar r_g(c)$ for 2 groups: $\mathcal{S}_{\text{gold human}}$ and $\mathcal{S}_{\text{gold machine}}$.
Intuitively, a value such as $\bar r_{\text{gold machine}}(c)=0.01$ means that, on average, \emph{1\%} of all retained patches in a machine-labeled sample belong to a cluster $c$; a difference of $0.15$ corresponds to a $15\%$ shift in the average patch share.

\paragraph{Motif discriminativeness across datasets and proxy models.}
Given gold-labeled groups $\mathcal{S}_{\text{gold human}}$ and $\mathcal{S}_{\text{gold machine}}$, we quantify how strongly a motif cluster separates classes via the difference in mean motif rates
\begin{equation}
\Delta r(c)\;=\;\bar r_{\text{gold machine}}(c)\;-\;\bar r_{\text{gold human}}(c).
\label{eq:delta-r}
\end{equation}
Using the aggregate statistics (top-3 clusters by $|\Delta r(c)|$ per setting), we find that the most discriminative motif often shifts by several percentage points in average patch share, and in some cases by more than $10$ points.
Table~\ref{tab:top1-motif} reports the top-1 cluster per dataset/proxy/preprocessing variant, including bootstrap confidence intervals for $\Delta r(c)$ and a permutation-test $p$-value.
In \textsc{HC3}, the strongest motif (cid=24) is \emph{machine-enriched}, increasing from $9.7\%$ (gold human) to $20.7\%$ (gold machine), $\Delta r=+11.0$ points with a tight $95\%$ CI $[9.3,12.7]$.
In contrast, for \textsc{RAID} the top motifs are consistently \emph{human-enriched} under all tested proxy models and both preprocessing variants (e.g., $\Delta r=-10.5$ points for \textsc{Cohere} under the unified variant).
Qualitatively, the corresponding example patches (Fig.~\ref{fig:motif-std-gpt-human}) show similar, repeatable local structures: for human-leaning patch clusters, ``islands'' and isolated flashes of attention are observed across all datasets and models, while the only identified heavy machine-inclined cluster of \textsc{HC3}/\textsc{GPT-neo} exhibits a prevalence of the horizontal bands. The notable outlier is GPT-neo for RAID, which is explained by HDBSCAN producing a very low amount of dense clusters (n = 4), resulting in a high intra-class variance. Such observations support the interpretation of clusters as visually coherent ``motif families''.

\begin{table*}[t]
\centering
\caption{Top-1 motif cluster by absolute class-rate gap $|\Delta r(c)|$ for each dataset/setting. Rates are mean per-sample motif shares (\%), and $\Delta r$ is reported in percentage points (machine minus human). Confidence intervals are 95\% bootstrap CIs over samples; $p$ is from a label-permutation test.}
\label{tab:top1-motif}
\setlength{\tabcolsep}{6pt}
\scriptsize
\renewcommand{\arraystretch}{1.4}
\begin{tabular}{lllc rrr r r l}
\toprule
\textbf{Dataset} & \textbf{Setting} & \textbf{Proxy-model} & \textbf{cid}
& \textbf{Human $\bar r_g$} & \textbf{Machine $\bar r_g$} & \textbf{$\Delta r$}
& \textbf{95\% CI} & \textbf{$p$} \\
\midrule
HC3  & individual     & \textsc{GPT-neo} 2.7B    & 24 &  9.7 & 20.7 & +11.0 & [ 9.3, 12.7] & $<\!0.001$ \\
RAID & unified & \textsc{Cohere} Cmd-R 7B  & 13 & 80.0 & 69.5 & -10.5 & [-12.9, -8.2] & $<\!0.001$ \\
RAID & individual     & \textsc{Mistral} 7B       & 23 & 63.5 & 54.8 &  -8.7 & [-11.1, -6.2] & $<\!0.001$ \\
RAID & individual     & \textsc{GPT-neo} 2.7B     &  4 & 87.6 & 81.1 &  -6.5 & [ -9.0, -4.1] & $<\!0.001$ \\
RAID & individual     & \textsc{Llama} 3.1 8B     & 12 &  7.5 &  3.1 &  -4.4 & [ -4.8, -3.9] & $<\!0.001$ \\
RAID & unified & \textsc{GPT-neo} 2.7B     & 37 & 11.6 &  8.5 &  -3.0 & [ -4.1, -2.0] & $<\!0.001$ \\
RAID & unified & \textsc{Mistral} 7B       & 39 &  7.3 &  5.1 &  -2.2 & [ -2.9, -1.6] & $<\!0.001$ \\
RAID & individual     & \textsc{Cohere} Cmd-R 7B  &  8 & 82.1 & 79.9 &  -2.1 & [ -3.7, -0.6] & 0.031 \\
RAID & unified & \textsc{Llama} 3.1 8B     & 16 &  9.5 &  7.6 &  -2.0 & [ -2.4, -1.5] & $<\!0.001$ \\
\bottomrule
\end{tabular}
\end{table*}

\paragraph{Patch-wise saliency and ablation are computed per retained patch.}
To relate motif \emph{prevalence} to motif \emph{importance}, we assign each retained patch an (i) Grad-CAM\citep{Selvaraju_2019} score and (ii) a zeroing-ablation score, and then aggregate these scores by cluster.
Concretely, for each retained patch index $(k,u,v)\in\mathcal{I}_s$ we compute:
(i) a Grad-CAM activation $g_{s,k,u,v}$ at the last convolutional stage (the same $16\times16$ grid that defines the patches), and
(ii) an ablation-induced logit change $\delta_{s,k,u,v}$ obtained by zeroing the corresponding $8\times8$ region in the input block $A_k$ and re-evaluating the detector.
We then summarize cluster-level importance by averaging over all occurrences assigned to cluster $c$:
\begin{equation}
\begin{aligned}
\bar g(c) &= \mathbb{E}\!\left[g_{s,k,u,v}\mid c_{k,u,v}=c\right],\\
\bar\delta(c) &= \mathbb{E}\!\left[\delta_{s,k,u,v}\mid c_{k,u,v}=c\right].
\end{aligned}
\end{equation}
Here $\bar\delta(c)$ captures the \emph{average marginal effect} of removing a single motif instance, while $\bar r_g(c)$ captures \emph{how frequently} the motif occurs in a given group.

\paragraph{Why prevalence need not correlate with saliency or ablation.}
Empirically, we observe that clusters with the largest $|\Delta r(c)|$ are not necessarily the clusters with the highest $\bar g(c)$ or $|\bar\delta(c)|$.
Aggregated correlations across all datasets and models are weak and unstable:
Across top-15 clusters for Grad-CAM
Pearson $\rho=\text{0.143}\pm0.23$,  Spearman $\rho=\text{0.10}\pm0.29$;
for Zero-ablation
Pearson $\rho=\text{-0.03}\pm0.36$,  Spearman $\rho=\text{-0.09}\pm0.25$.
Across the top-3 clusters, correlation becomes moderate for GRAD-CAM but still unstable:
Pearson $\rho=\text{0.373}\pm0.66$,  Spearman $\rho=\text{0.37}\pm0.54$;
while it slightly improves for Zero-ablation,
Pearson $\rho=\text{0.09}\pm0.67$,  Spearman $\rho=\text{0.12}\pm0.64$.
We attribute this mismatch to the following reasons:
First, a dataset imbalance can decouple prevalence from importance. Taking into account the significant imbalance of used datasets, the RAID subset and HC3, training under skewed prior biases the detector toward features that optimize empirical risk for the majority class, so a motif may be strongly pronounced in $\Delta r(c)$ yet have muted average Grad-CAM/ablation effects.
Second, for Grad-CAM and zeroing-ablation the same motif family may be decisive only in certain positions or alongside other motifs; averaging $\bar g(c)$ and $\bar\delta(c)$ across all occurrences can weaken these conditional effects.

Taken together, these findings suggest that motif analysis is best interpreted as a complementary view:
$\Delta r(c)$ reveals dataset and proxy-model dependent differences in local attention map structure,
while $\bar g(c)$ and $\bar\delta(c)$ reflect how the trained detector \emph{uses} (or ignores) individual motif instances at decision time conditioned on the training data.

\paragraph{Implications.}
Overall, the presence of statistically reliable shifts in patch motif rates between gold machine and gold human groups (Table~\ref{tab:top1-motif}) supports our hypothesis that the internal dynamics of a proxy $G_\theta$ induce detectable structure in attention-based attribution maps beyond surface-level text statistics.
At the same time, the weak alignment between prevalence and saliency cautions against interpreting frequent motifs as predictions of the model behaviour, instead, they appear to function as stable, repeatable signatures that the detector can exploit in combination with other cues.

\section{Conclusion}

We presented \textsc{AEyeDE}, an attribution-based framework for AI-generated text detection that leverages attention-derived attribution maps from a proxy Transformer as a signal to learn a model to distinguish between human-written and AI-generated text. Across both encoder-decoder translation benchmarks (WMT14, UN) and decoder-only generation datasets (\textsc{HC3}, \textsc{RAID}), as well as adversarial attacks and unseen cross-dataset samples, \textsc{AEyeDE} achieves competitive performance and shows strong results. This is achieved by training a lightweight CNN model on a relatively small amount of data.

Beyond evaluation metrics, we provide an analysis of localized attribution patterns and show that they systematically differ between human and AI-generated text. Overall, our results suggest that internal attribution behavior offers a complementary and effective signal for reliable authorship detection, motivating further work on broader robustness settings and alternative attribution sources.

\paragraph{Limitations}

Our study has some limitations. First, the proposed framework assumes \textit{white-box} access to a proxy Transformer model in order to extract attention-based attribution maps. While this assumption may limit applicability in fully black-box settings, our experiments indicate that attribution patterns generalize across generator families, suggesting that exact access to the true generator is not strictly required.

Furthermore, our implementation focuses on attention-based attributions, which offer a favorable trade-off between informativeness and computational cost for large models and long sequences. Investigating alternative attribution methods, such as gradient-based saliency, may further enrich the analysis, but is left for future work due to their higher computational overhead. Additionally, we computed attention attribution by averaging over all layers and heads. Investigating a more fine-grained selection strategy for this task remains for future work.

\bibliography{tmlr}
\bibliographystyle{tmlr}

\appendix

\end{document}